\documentclass[3p]{elsarticle}

\usepackage{hyperref}
\usepackage{subfigure}
\usepackage{amsmath}
\usepackage{graphicx}
\usepackage{color}
\usepackage{colortbl}
\graphicspath{{./images/}}

\makeatletter
\renewcommand*\env@matrix[1][c]{\hskip -\arraycolsep
  \let\@ifnextchar\new@ifnextchar
  \array{*\c@MaxMatrixCols #1}}
\makeatother

\begin{document}

\begin{frontmatter}

    \title{Serial-EMD: Fast Empirical Mode Decomposition Method for Multi-dimensional Signals Based on Serialization}
    
    \author[1]{Jin Zhang \fnref{fn1}}
    \ead{nkzhangjin@nankai.edu.cn}
    \author[2]{Fan Feng \fnref{fn1}}
    \ead{sir_fengfan@163.com}
    \author[3]{Pere Marti-Puig}
    \ead{pere.marti@uvic.cat}
    \author[2,4,5]{Cesar F. Caiafa}
    \ead{ccaiafa@gmail.com}
    \author[6]{Zhe Sun}
    \ead{zhe.sun.vk@riken.jp}
    \author[2]{Feng Duan}
    \ead{duanf@nankai.edu.cn}
    \author[2,3]{Jordi Solé-Casals\corref{cor1}}
    \ead{jordi.sole@uvic.cat}

    \cortext[cor1]{Corresponding author}
    \fntext[fn1]{These authors contributed equally to this work.}

    \address[1]{College of Computer Science, Nankai University,
    300071 Tianjin, China}
    \address[2]{College of Artificial Intelligence, Nankai University, 300350 Tianjin, China}
    \address[3]{Data and Signal Processing Group, University of Vic—Central University of Catalonia,
    08500 Vic, Catalonia, Spain}
    \address[4]{Instituto Argentino de Radioastronom\'{i}a, CONICET CCT La Plata/ CIC-PBA / UNLP, 1894 V. Elisa, Argentina}
    \address[5]{Tensor Learning Team  - Center for Advanced Intelligence Project, RIKEN, 103-0027 Tokyo, Japan}
    \address[6]{Computational Engineering Applications Unit, Head Office for Information Systems and Cybersecurity, RIKEN, 351-0198 Saitama, Japan}

    \begin{abstract}
        Empirical mode decomposition (EMD) has developed into a prominent tool for adaptive, scale-based signal analysis in various fields like robotics, security and biomedical engineering. Since the %dimension explosion of massive data
        dramatic increase in amount of data puts forward higher requirements for the capability of real-time signal analysis, it is difficult for existing EMD and its variants to trade off the growth of data dimension and the speed of signal analysis. In order to decompose multi-dimensional signals at a faster speed, we present a novel signal-serialization method (serial-EMD), which concatenates multi-variate or multi-dimensional signals into a one-dimensional signal and uses various one-dimensional EMD algorithms to decompose it. To verify the effects of the proposed method, synthetic multi-variate time series, artificial 2D images with various textures and real-world facial images are tested. Compared with existing multi-EMD algorithms, the decomposition time becomes significantly reduced. In addition, the results of facial recognition with Intrinsic Mode Functions (IMFs) extracted using our method can achieve a higher accuracy than those obtained by existing multi-EMD algorithms, which demonstrates the superior performance of our method in terms of the quality of IMFs. Furthermore, this method can provide a new perspective to optimize the existing EMD algorithms, that is, transforming the structure of the input signal rather than being constrained by developing envelope computation techniques or signal decomposition methods. In summary, the study suggests that the serial-EMD technique is a highly competitive and fast alternative for multi-dimensional signal analysis.
        
    \end{abstract}

    \begin{keyword}
        Empirical mode decomposition \sep Signal serialization
    \end{keyword}

\end{frontmatter}
% \linenumbers
\section{Introduction}

% Introduce EMD

Empirical mode decomposition (EMD) \cite{Huang1998} is a versatile data-driven method that quantitatively decomposes real-world signals, whereby the original signal is modeled as a linear combination of amplitude-modulated and frequency-modulated (AM-FM) functions, called intrinsic mode functions (IMFs) \cite{Sharpley2006}. Together with the Hilbert transform \cite{Tolwinski2007}, the IMFs yield meaningful instantaneous frequency estimates that give sharp identifications of imbedded structures. The complete process is the so-called Hilbert-Huang Transform (HHT) \cite{Huang2005}. %Owing to no a priori hypothesis regarding the nature of the data, 
With no prior assumption on the nature of the data, EMD and its noise-assisted variants, such as ensemble EMD (EEMD) \cite{Wu2009} and complete ensemble EMD with adaptive noise (CEEMDAN) \cite{Torres2011}, %have been applied in numerous applications in diverse fields,
have been put to use in numerous applications in diverse fields, such as in biomedical engineering \cite{Gallix2012}, functional neuroimaging \cite{Navarro2012}, image enhancement \cite{Celebi2012} or fault diagnosis of mechanical systems \cite{Lei2011}. However, these extensions are limited in applicability to %%signal channel (uni-variate) time series (one-dimensional),
one-dimensional time series, which was a major incentive for the numerous variations of the standard EMD algorithm.

Recent advances in physics and engineering %have brought to light new problems dealing with
have brought new problems to light which deal with complexity, uncertainty, non-linearity and multi-channel dynamics \cite{UmairBinAltaf2007}. Over the past decade, EMD has been successfully extended to be capable of handling multi-variate, multi-dimensional signal decomposition. These kinds of extensions can be collectively called multi-EMD algorithms. For multi-dimensional signals, extracting local extrema to estimate the envelopes (which is the key issue in EMD algorithms) is not as straightforward as for one-dimensional signals, because the complex and quaternion fields are not ordered \cite{Mandic2009}. The first attempts at a multi-variate EMD \cite{Tanaka2007, Rilling2007, UrRehman2010} were restricted to bivariate and trivariate cases, and were later extended to n-variate signals by means of the Multi-Variate EMD (MVEMD) \cite{Rehman2010}. These extensions use various projection techniques to help capture the envelopes, where the data channels are projected onto numerous direction vectors. On the other hand, the primitive multi-dimensional capabilities of EMD were also restricted to two-dimensional data (images) \cite{Rilling2007}. Pseudo-bidimensional EMD (pseudo BEMD) \cite{Wu2009a}, the first approach, performed one-dimensional EMD on successive slices of 2D data, but generated poor IMF components due to the fact that it ignored the correlation among the rows and/or columns of a 2D image. More appropriate bidimensional EMD algorithms, such as bidimensional EMD (BEMD) \cite{Nunes2003} and image EMD (IEMD) \cite{Linderhed2009}, employed more complicated interpolation methods to estimate the envelopes, such as radial basis functions and thin plate splines rather than the cubic spline interpolation of the standard EMD.

However, in multi-EMD it is generally required to find local extrema to interpolate these points in each iteration of the process. These kinds of interpolation schemes have to face the following issues:

\textbf{Dimensional expansion:} In a standard one-dimensional EMD, the IMFs must satisfy two criteria: (1) the number of extrema and the number of zero-crossings must either be equal or differ at most by one; and (2) the mean value of the envelope defined by the local maxima and local minima must be zero. The situation is more difficult in the case of a multi-EMD, as parameters required for defining an IMF, such as zero-crossings, are not so well defined in more than one dimension.

\textbf{Computational load:} %Whether projection techniques in multi-variate EMD algorithms or interpolation methods in multi-dimensional EMD algorithms, these morphological operations \cite{Huang1996} 
Morphological operations \cite{Huang1996}, such as projection techniques in multi-variate EMD algorithms or interpolation methods in multi-dimensional EMD algorithms, render the envelope identification process complicated and time consuming, as it requires finding an interpolation surface on high dimensional data space during each iteration. The computational load of such algorithms is prohibitive for large-scale and complex scenarios.

%Focusing on optimizing the current poor envelope determination methods across expansive domains, a few EMD extensions have been proposed to overcome the large computational cost requirements. 
A few EMD extensions, focusing on optimizing the current poor envelope determination across expansive domains, have been proposed to overcome the large computational load requirements. Fast and adaptive BEMD (FABEMD) \cite{Bhuiyan2010} employed order statistic filters (OSFs) to get the upper and lower envelopes in the decomposition process instead of using surface interpolation. Green's function in tension BEMD (GiT-BEMD) \cite{Al-Baddai2016} introduced the Green's function as a surface interpolation technique, which in addition needed very few iterations to estimate each IMF. Most of these modifications have been specifically designed for bidimensional signals. However, general fast-speed n-variate or multi-EMD algorithms are still in their infancy, and further efforts are needed to speed up the signal decomposition.

%In this study, we propose, for the first time, a new simple signal serialization technique, 
In this study, we propose a simple and novel signal serialization technique, %thereby concatenating multi-variate or multi-dimensional signals into a one-dimensional signal
through which we can concatenate multi-variate or multi-dimensional signals into a one-dimensional signal. This paves the way for performing multi-signal decomposition through one-dimensional EMD algorithms. Most favourably, this concatenation method can obviously reduce the computational time without altering the existing EMD algorithms. Furthermore, it can provide a new perspective to optimize current EMD algorithms, that is, transforming the structure of input signals instead of developing new envelope identification techniques or new multi-EMD algorithms. Thus, the proposed fast serial-EMD method can be a good alternative, providing efficient signal analysis and mode decomposition.

The rest of the paper is organized as follows: before introducing the concept of serial-EMD, the standard EMD and its variants will be briefly outlined in Section \ref{sec_02}. The description of the proposed serial-EMD algorithm will be discussed in Section \ref{sec_03}. Results and discussion on various EMD algorithms with simulated and real-world datasets will be given in Section \ref{sec_04}. Finally, a conclusion will be drawn in Section \ref{sec_05}.
\section{The classical EMD and its variants}
\label{sec_02}

The EMD and its variants can break down a signal into a finite sum of components. These components form a complete and nearly orthogonal basis for the original signal, which can be described in terms of IMFs.

\subsection{Standard one-dimensional EMD}

For a real-valued signal $x(k)$, the one-dimensional standard EMD \cite{Huang1998} finds a set of $N$ IMFs $\{c_i(k)\}^N_{i=1}$, and a residue signal $r(k)$, so that

\begin{equation}
    x(k) = \sum_{i=1}^N c_i(k) + r(k)
\end{equation}

IMFs $c_i(k)$ are defined so as to have symmetric upper and lower envelopes, with the number of zero crossings and the number of extrema differing at most by one. The standard one-dimensional EMD algorithm can be described as follows:

\begin{enumerate}[Step 1.]
    \item Find the locations of all the extrema of $x(k)$.
    \item Interpolate between extrema to obtain the lower (upper) signal envelope $e_{min}(k)$ ($e_{max}(k)$).
    \item Compute the local mean $m(k) = [e_{min}(k) + e_{max}(k)] / 2$.
    \item Compute the IMF candidate $s(k) = x(k) - m(k)$.
    \item If $s(k)$ satisfies some predefined stopping criterion, then define $c_i(k) = s(k)$ as an IMF, otherwise set $x(k) = s(k)$ and repeat the process from Step 1.
\end{enumerate}

The refinement process (steps 2 to 5), which is needed to extract every mode, requires a certain number of iterations and is named as the sifting process.

\subsection{Noise-assisted EMD}

One of the major drawbacks of the standard one-dimensional EMD is the frequent appearance of mode mixing \cite{Hu2012}, which is defined as either a single IMF consisting of signals of widely disparate scales, or a signal of a similar scale residing in different IMF components. The ensemble EMD (EEMD) \cite{Wu2009} was proposed to counteract the problem; it performs the decomposition over an ensemble of noisy copies of the original signal and obtains the final results by averaging. The EEMD algorithm can be described as follows:

\begin{enumerate}[Step 1.]
    \item Generate $x^{(m)} = x + \beta w^{(m)}$, where $w^{(m)}$ $(m=1,...,M)$ are different realizations of white noise.
    \item Decompose each $x^{(m)}$ $(m=1,...,M)$ completely by EMD, obtaining its modes ${c}_i^{(m)}$, where $i=1,...N$ indicates the modes.
    \item Assign $\overline{c_i}$ as the $i$th mode of $x$, obtained by averaging the corresponding modes.
\end{enumerate}

It should be noticed that in EEMD, every $x^{(m)}$ is decomposed independently from different realizations and for every one of them a residue $r_i(k) = x(k) - c_i(k)$ is obtained at each stage. The extraction of every $x^{(m)}$ $(m=1,...,M)$ requires a different number of sifting iterations. This situation can cause different realizations of signal plus noise, might produce different number of modes and residual noise might be involved in the modes. %Taking into account these drawbacks,
Taking these drawbacks into account, the complete ensemble EMD with adaptive noise (CEEMDAN) \cite{Torres2011} was proposed. The general idea is the following:

\begin{enumerate}[Step 1.]
    \item Compute the first residue $r_1 = x - \widetilde{c_1}$, where $\widetilde{c_1}$ is the first CEEMDAN mode, namely, the first mode obtained by EEMD.
    % \item Obtain the first mode of $r_1 + \beta_1 E_1 (w^{(m)})$ by EMD, $m=1,...,M$, and compute the second CEEMDAN mode $\widetilde{c_2}$ by averaging these first mode.
    \item Obtain each first mode of $r_1 + \beta_1 E_1 (w^{(m)})$ by EMD, $m=1,...,M$, and compute the second CEEMDAN mode $\widetilde{c_2}$ by averaging these first modes. %IN HERE, ARE YOU REFERRING TO AVERAGING THE FIRST MODE OR BY AVERAGING VARIOUS FIRST MODES (YOU USED "THESE", SUGGESTING A PLURAL IN FIRST MODE, BUT I'M UNSURE)
    %%%%% YES, THAT SHOULD BE MULTIPLE FIRST MODES.
    \item For $i=2,...,N$, calculate the $i$th residue $r_i = r_{(i-1)} - \widetilde{c_i}$.
    \item Obtain each first mode of $r_i + \beta_i E_i (w^{(m)})$ by EMD, $m=1,...,M$, and compute the $i$th CEEMDAN mode $\widetilde{c_2}$ by averaging these first modes %IN HERE, ARE YOU REFERRING TO AVERAGING THE FIRST MODE OR BY AVERAGING VARIOUS FIRST MODES (YOU USED "THESE", SUGGESTING A PLURAL IN FIRST MODE, BUT I'M UNSURE).
    %%%%% SAME AS THE PREVIOUS ONE.
\end{enumerate}

\subsection{Multivariate EMD}

The general approach for $n$-variate signals, multivariate EMD (MVEMD) \cite{Rehman2010}, has the ability to extract common modes across the signal components, making it suitable for the fusion of information from multiple source. The MVEMD algorithm can be summarized as follows:

\begin{enumerate}[Step 1.]
    \item Choose a suitable point set for sampling on an $(n-1)$ sphere.
    \item Calculate a projection, denoted by $\{p^{\theta_j}(k)\}_{k=1}^K$, of the input signal $\{\mathbf{v}(k)\}_{k=1}^K$ along the direction vector $\mathbf{x}^{\theta_j}$, for all $j$ (the whole set of direction vectors), giving $\{p^{\theta_j}(k)\}_{j=1}^J$ as the set of projections.
    \item Find the time instants $\{t_i^{\theta_j}\}$ corresponding to the maxima of the set of projected signals $\{p^{\theta_j}(k)\}_{j=1}^J$.
    \item Interpolate $[t_i^{\theta_j}, \mathbf{v}(t_i^{\theta_j})]$ to obtain multivariate envelope curves $\{\mathbf{e}^{\theta_j}(k)\}_{j=1}^J$.
    \item For a set of $J$ direction vectors, calculate the mean $\mathbf{m}(k)$ of the envelope curves by averaging $\mathbf{e}^{\theta_j}(k)$.
    \item Extract the detail $d(k)$ using $d(k)=x(k)-m(k)$. If the detail $d(k)$ fulfills the stoppage criterion for a multivariate IMF, apply the above procedure to $x(k)-d(k)$, otherwise apply it to $d(k)$.
\end{enumerate}

\subsection{Multidimensional EMD}

The bidimensional EMD (BEMD) \cite{Nunes2003} is one of the most famous two-dimensional EMD algorithms on 2D image processing, and its variants \cite{Damerval2005, Nunes2009, Liu2005} with various interpolation methods have been thoroughly studied. A typical sifting process for a general BEMD can be described as follows:

\begin{enumerate}[Step 1.]
    \item Set $r(x, y) = X(x, y)$. Identify all local maxima and local minima of $r$.
    \item Interpolate the local minima (maxima) to obtain the lower (upper) envelope surface $e_{min}(x, y)$ ($e_{max}(x, y)$).
    \item Calculate the mean of the envelope surface $m(x, y)$ and subtract it from $r(x, y)$ to update $r(x, y)$.
    \item Repeat Step 1 to 3 until the stopping criterion is met.
\end{enumerate}

The state-of-the-art BEMD variant, Green's function in tension BEMD (GiT-BEMD) \cite{Al-Baddai2016}, which replaces the direct spline interpolation step by an interpolation based on Green's function \cite{Wessel1998}, turned out to be very efficient in practical applications such as image analysis \cite{Al-Baddai2020, Yang2016}. Interpolation using Green's function implies that the points of the interpolating envelope surface can be expressed as

\begin{equation}\label{eq:green_function}
    s(\mathbf{x}_u) = \sum_{n=1}^N w_n \phi(\mathbf{x}_u, \mathbf{x}_n)
\end{equation}

where $\mathbf{x}_u$ denotes any point where the surface is unknown, $\mathbf{x}_n$ presents the $n$th data constraint, $\phi(\mathbf{x}_u, \mathbf{x}_n)$ is the Green's function and $w_n$ is the respective weight in the envelope representation. The GiT-BEMD can be summarized in three steps: 
%HERE YOU'RE TALKING ABOUT 2 STEPS, BUT THEN THERE'S ACTUALLY 3 STEPS. I ASUSME YOU MEANT TO WRITE "THE GiT-BEMD CAN BE SUMMARIZED IN THREE STEPS"?
%%%%% YES, THAT SHOULD BE "THREE STEPS"

\begin{enumerate}[Step 1.]
    \item Calculate the local extrema $\mathbf{s}(\mathbf{x}_n)$ by using an 8-connected neighbourhood strategy, and estimate the weights $\mathbf{w}$ using eq. \ref{eq:green_function} with locations $\mathbf{x}_n = {[x_n, y_n]}^T$.
    \item %PLEASE CAN YOU CHECK THAT THIS STEP (2) MAKES SENSE?
    %%%%% THAT MAKES SENSE, I JUST REF THE ORIGINAL GIT-BEMD PAPER.
    Using the interpolation eq. \ref{eq:green_function} for each of the known points, a linear system with $N$ equations is obtained. Then, solve for the weights $\mathbf{w}=\mathbf{G^{-1}c}$.
    \item Using the weights $\mathbf{w}$, the value $s(\mathbf{x}_u) \equiv c_u$ of the envelope surface can be estimated at any point $\mathbf{x}_u$ by solving eq. \ref{eq:green_function}, where the vector $\mathbf{\phi} = {[\phi(\mathbf{x}_u, \mathbf{x}_1), \phi(\mathbf{x}_u, \mathbf{x}_2) ... \phi(\mathbf{x}_u, \mathbf{x}_N)]}^T$ contains the Green's function values of all distances between the $N$ data constraints and the considered location.
\end{enumerate}

Various complicated envelope estimation techniques during each sifting iteration can be used to characterize these EMD extensions. The poor performance and huge difficulty when interpolating points across expansive domains are the primary reason for the heavy computational burden of multi-EMD algorithms. In this paper, the mentioned EMD algorithms, EMD, EEMD, CEEMDAN, MVEMD and GiT-BEMD will be used for performance comparison with the proposed serial-EMD.
\section{The proposed serial-EMD algorithm}
\label{sec_03}

With the aim of reducing the computational load during the sifting process for multivariate or multi-dimensional EMD algorithms, a novel serial-EMD approach is proposed here based on a serialization technique. %Firstly,  a  multi-dimensional  signal  can  be  treated  as  composed  by  multiple  vectors(one-dimensional  signals)
To start with, note that a multi-dimensional signal can be treated as being composed by multiple vectors (one-dimensional signals). Then, using a transition, which is calculated by using some part of the data from each dimension, the multi-dimensional signal can be serialized. After that, the serialized one-dimensional signal is decomposed by a standard one-dimensional EMD or its variants and IMFs for serialized signal are calculated. Finally, after removing the redundant transitions, the IMFs of the original multi-signals are extracted from the IMFs from the serialized signal by reshaping and slicing. The details of the signal concatenation and IMF deconcatenation techniques are discussed in the following subsection. 

\subsection{Serial concatenation of two signals}

\begin{figure}[ht]
    \centering{}
    \includegraphics[width=0.8\textwidth]{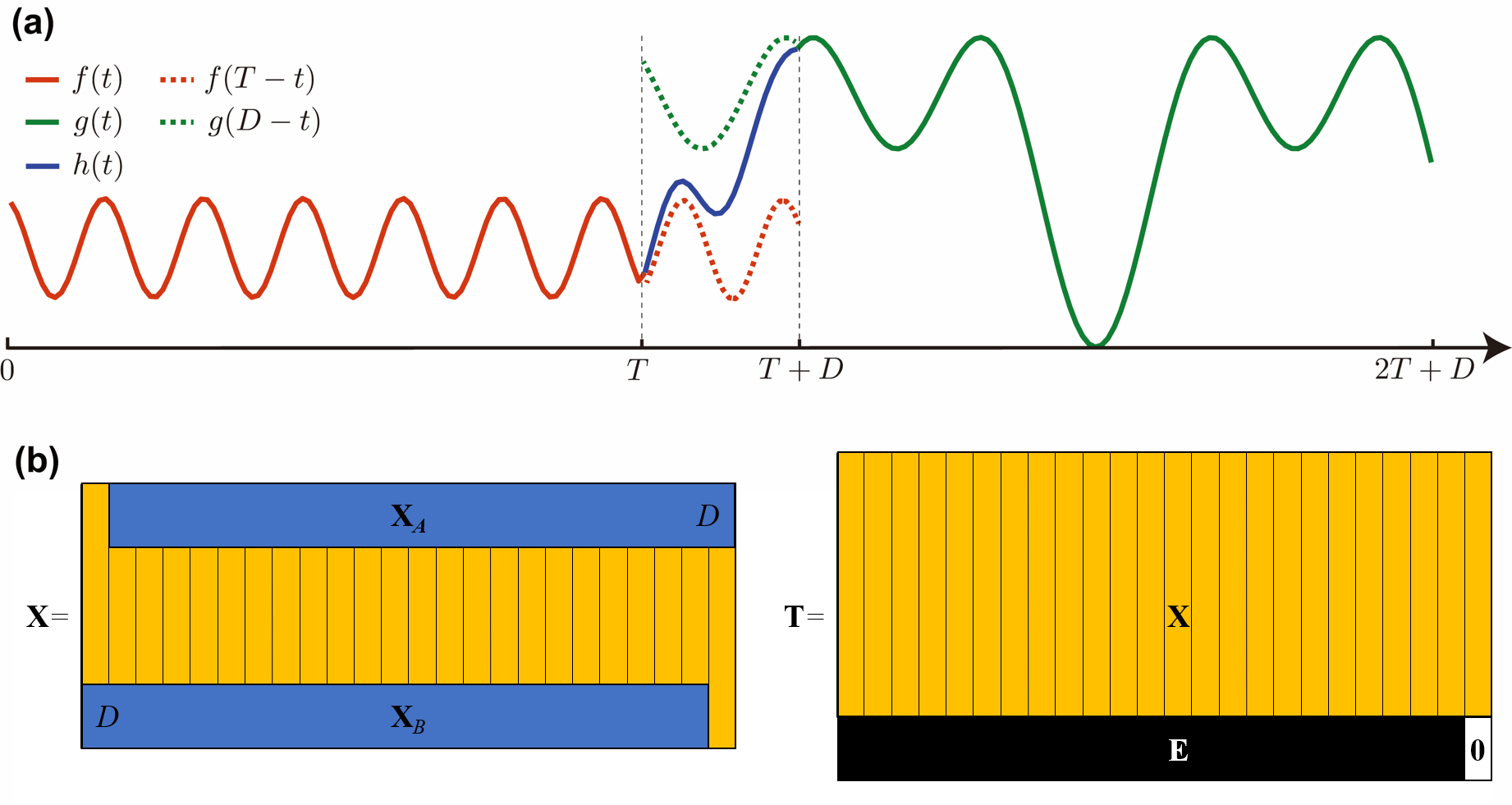}
    \caption{(a) The concatenation of signals $f(t)$ and $g(t)$ by inserting a transition signal $h(t)$ computed through eq. (\ref{eq:transition}). The concatenated signal maintains the continuity and characteristics of the original signals. The only parameter we need to choose is the length of transition $D$. (b) Details of the concatenation algorithm when applied to multiple signals arranged as columns of matrix $\mathbf{X}$. The final result is stored in matrix $\mathbf{T}$.}
    \label{fig:C2}
\end{figure}

Given two continuous functions $f(t)$ and $g(t)$ with $t\in [0,T]$, let us first consider the problem of how to concatenate those signals avoiding a discontinuity in the joint, i.e. assuming that in general $f(T)\neq g(0)$.

We propose to define a transition function $h(t)$ of length $D<T$, with $t\in [0,D]$, which is to be inserted in between $f(t)$ and $g(t)$ such that the resulting function $s(t)$, with $t\in [0,2T+D]$, is continuous. Specifically, $s(t)$ is defined as the serial concatenation of $f(t)$, $h(t)$ and $g(t)$ as follows:

\begin{equation}
    s(t) =  \left\{
    \begin{array}{ll}
        f(t)     & t\in[0,T]      \\
        h(t-T)   & t\in[T,T+D]    \\
        g(t-T-D) & t\in[T+D,2T+D]
    \end{array}
    \right.
\end{equation}

A simple way of obtaining such a transition function $h(t)$ is the following:

\begin{equation}\label{eq:transition}
    h(t) = \left(1 - \frac{t}{D}\right)f(T-t) + \frac{t}{D}g(D-t),
\end{equation}

\noindent with $t\in[0,D]$. It is easy to show that by using this transition function $h(t)$ the continuity of the resulting function $s(t)$ is guaranteed because $h(0)=f(T)$ and $h(D)=g(0)$. It is noted that $f(T-t)$ and $g(D-t)$, with $t\in[0,D]$, correspond to the flipped versions of the tail of $f(t)$ and the beginning part of $g(t)$ with lengths $D$. Figure \ref{fig:C2}(a) shows the effect of the proposed continuous concatenation of two signals.

\subsection{The fast multiple serial concatenation algorithm}

Let's consider $N$ signals of length $M$ organized as columns $\in \mathbf{R}^M$ $\mathbf{x}_i$, ($i=1, \dots, N$) of matrix $\mathbf{X}$ of size $ M \times N $ as follows:

\begin{equation}
    \mathbf{X}_{(M \times N)}=\begin{bmatrix} \mathbf{x}_1 & \mathbf{x}_2 & \cdots & \mathbf{x}_N
    \end{bmatrix}
\end{equation}

Let $D$ be the number of points we add between the signals to define the transitions in their concatenation. Then, consider the vector $\mathbf{a}$ with elements $a_i=i/(D+1)$, $i=1, \dots, D$, of evenly spaced values between 0 and 1, but with both values not included. In order to compute the set of filling points, we define the sub-matrices of $\mathbf{X}$, $\mathbf{X_A}$ and $\mathbf{X_B}$, both of size $D \times N-1$ in terms of the indices of $\mathbf{X}$ as follows (see Figure \ref{fig:C2}(b)-left):

\begin{equation}
    \mathbf{X_A}_{(D \times N-1)}=\mathbf{X}_{1:D \mathbf{,} 2:N}
\end{equation}

\begin{equation}
    \mathbf{X_B}_{(D \times N-1)}=\mathbf{X}_{M-D+1:M \mathbf{,} 1:N-1}
\end{equation}

Then we compute the signal extensions in a matrix $\mathbf{E}$ as follows:

\begin{equation}
    \mathbf{E}_{(D \times N-1)} =
    \mathbf{X_A}^f \odot \mathbf{a} \mathbf{u}^T + \mathbf{X_B}^f \odot \mathbf{a}^f \mathbf{u}^T
\end{equation}

\noindent where $\mathbf{u}=(1,...,1)$ of dimension $(N-1)$. The symbol $\odot$
stands for the the Hadamard product. The super-index $T$ means transposition and the super index $f$ means the flip up-to-down permutation, which flipped the order of the elements upside down along the first dimension. Note that the latter operation is obtained from pre-multiplying the matrix/vector affected by the standard involutory permutation, also known as backward identity or exchange matrix. The $f$ operation, however, can be done by appropriately addressing the elements, thus avoiding multiplications.

 %By using the length $D$ vector $\mathbf{z}$ of all-zero elements
 By using the zero-vector $\mathbf{z}$ of dimension $D$, we build the matrix $\mathbf{T}$ which contains the transitions introduced. Matrix $\mathbf{T}$ is defined in blocks as (see Figure \ref{fig:C2}(b)-right):

\begin{equation}
    \mathbf{T}_{(M+D \times N)} = \begin{bmatrix}[l]
        \mathbf{X}_{(M \times N)}                      \\
        \mathbf{E}_{(D \times N-1)} & \mathbf{z}_{(D)}
    \end{bmatrix}
\end{equation}

The vectorization of the matrix $\mathbf{T}$ provides the concatenated signals %through the transitions' added portions.
through the added portions of the transitions.

\begin{equation}
    \mathbf{t}_{((M+D) \cdot N)} = \operatorname{vec} ({\mathbf{T}})
\end{equation}

Finally, we eliminate the $\mathbf{z}$ vector %IF YOU ARE REFERRING TO ELIMINATING THE $\mathbf{z}$ VECTOR, YOU COULD WRITE THIS MORE CLEARLY AS:"WE ELIMINATE THE $\mathbf{z}$ VECTOR" INSTEAD OF "WE ELIMINATE THE $D$ ZEROS" 
%%%%% OK I AGREED WITH THIS.
introduced in $\mathbf{T}$ to obtain the useful samples in $\mathbf{x}$.

\begin{equation}
    \mathbf{x}_{(MN+DN-D)} = \mathbf{t}_{1:(M+D) \cdot N-D}
\end{equation}

Figure \ref{fig:C2} shows graphically an example of the concatenation of two consecutive signals, $\mathbf{x}_i$ and $\mathbf{x}_{i+1}$, through the $D$ filling samples $\mathbf{x_E}_{i}$ in terms of the elements involved in their formation. Note that using the vector $\mathbf{a}$, previously defined, and the $i$ column vectors $\mathbf{x_A}_i$, $\mathbf{x_B}_i$ of matrices $\mathbf{X_A}$, $\mathbf{X_B}$ respectively, the padding samples of column vector $\mathbf{x_E}_{i}$ of $\mathbf{X_E}$ can be written as: $\mathbf{x_E}_i = \mathbf{x_A}_i^f \odot \mathbf{a} + \mathbf{x_B}_i^f \odot \mathbf{a}^f$.

\subsection{The fast IMFs deconcatenation algorithm}

Let $F(\cdot)$ be any one-dimensional EMD algorithm or its variants, for the concatenated signals $\mathbf{x}$ without $D$ zeros attached. Its resulting IMFs $\mathbf{R}$ can be computed as follows:

\begin{equation}
    \mathbf{R}_{(MN+DN-D \times K)} = F(\mathbf{x}_{(MN+DN-D)})
\end{equation}

\noindent where $K$ is the number of IMFs. According to the definition of the EMD algorithm, the length of each IMF should be the same as the length of the signal to be decomposed. Thus, the transition part $\mathbf{E}$ in the concatenated signal $\mathbf{x}$ is also decomposed and included in the resulting IMFs $\mathbf{R}$.

\begin{figure}[ht]
    \centering{}
    \includegraphics[width=0.8\textwidth]{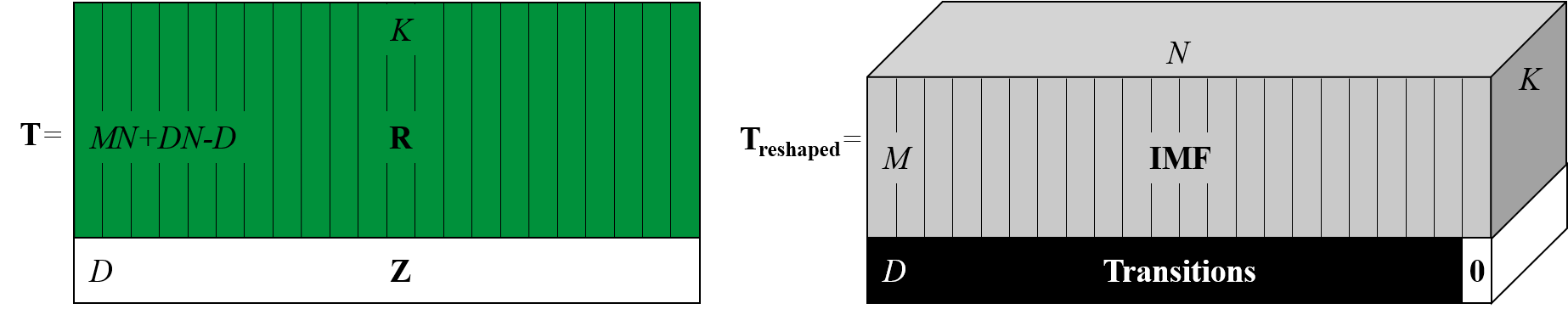}
    \caption{Detail of the deconcatenation algorithm applied to the resulting matrix $\mathbf{R}$ computed through EMD and its variants. The resulting matrix $\mathbf{R}$ can be filled with $D$ zeros for each column and can be reshaped to exclude the transitions. Notice that the matrix $\mathbf{Z}$ in the matrix $\mathbf{T}$ becomes the last column in the transitions after the reshaping operation. The pure IMFs of signals are stored in the matrix $\mathbf{IMF}$.}
    \label{fig:C3}
\end{figure}

In order to get the pure IMFs of every single signal, we build the matrix $\mathbf{T}$ to exclude the transitions of IMFs from the resulting IMFs $\mathbf{R}$, by using the matrix $\mathbf{Z}$ of size $D \times K$ of all-zero elements. Matrix $\mathbf{T}$ is defined in blocs as (see Figure \ref{fig:C3}-left):

\begin{equation}
    \mathbf{T}_{(MN+DN \times K)} = \begin{bmatrix}[l]
        \mathbf{R}_{(MN+DN-D \times K)} \\
        \mathbf{Z}_{(D \times K)}
    \end{bmatrix}
\end{equation}

Finally, matrix $\mathbf{T}$ can be reshaped into a new $(M+D) \times N \times K$ shape (see Figure \ref{fig:C3}-right). The IMFs of transitions in the concatenated signals are stored in the last $D$ columns after the reshaping operation. Thus, we can eliminate the $D \times N \times K$ transitions part introduced in $\mathbf{T}$ to get the pure IMFs.

\begin{equation}
    \mathbf{IMF}_{(M \times N \times K)} = reshape(\mathbf{T}, [M, N, K])
\end{equation}

The graphical description of the deconcatenation of IMFs can be found in Figure \ref{fig:C3}. Note that for the $i$th signal, its $j$th IMF can be written as: $\mathbf{imf}_{ij}=\mathbf{IMF}_{1:M \mathbf{,} i \mathbf{,} j}$.

\begin{figure}[ht]
    \centering{}
    \includegraphics[width=1.0\textwidth]{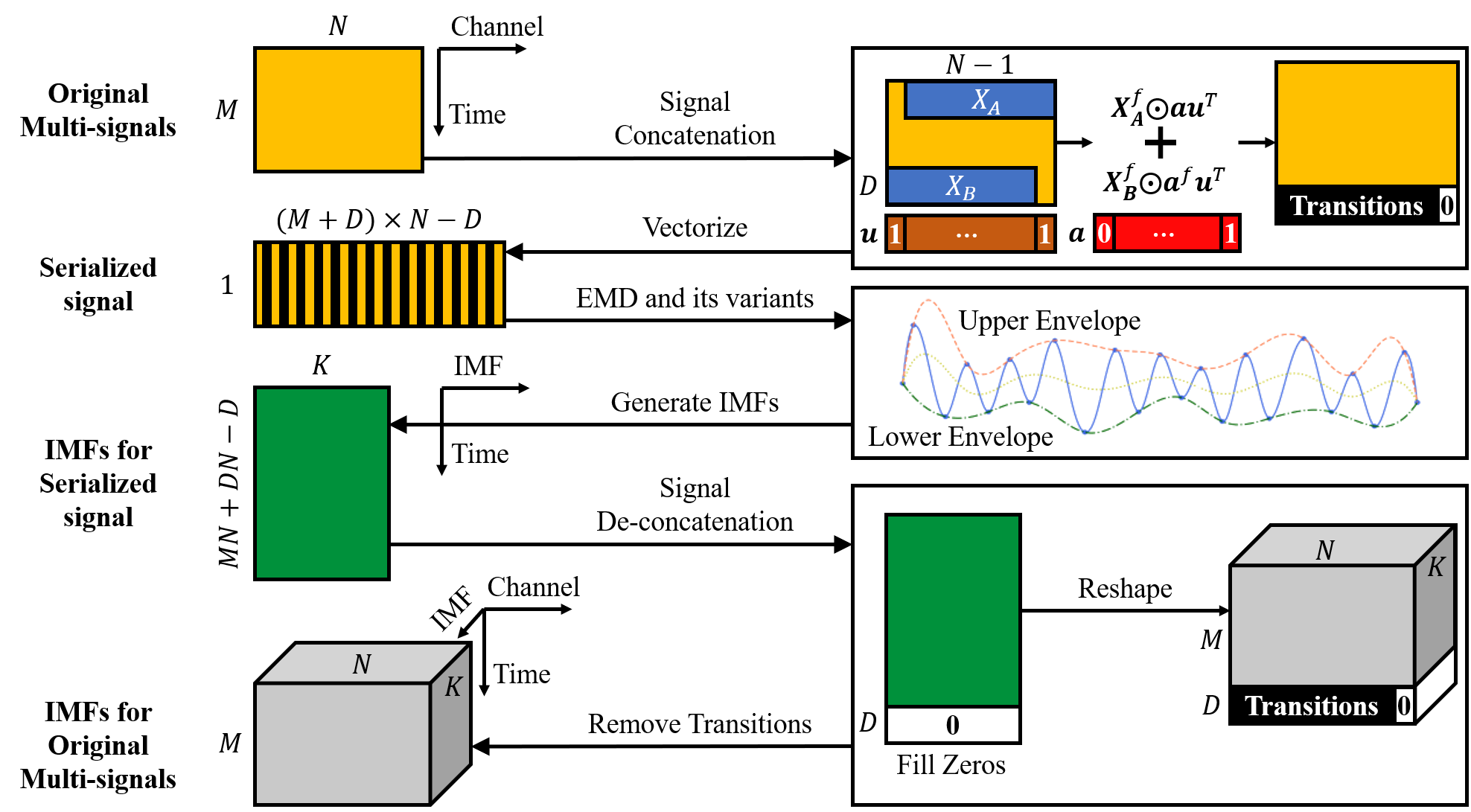}
    \caption{The description of the serial-EMD algorithm for 2D signals. The original multi-dimensional signals are concatenated by adding transitions. The IMFs for original multi-dimensional signals are extracted by removing the transitions.}
    \label{fig:C1}
\end{figure}

In summary, Figure \ref{fig:C1} explicitly outlines the steps involved in the serial-EMD algorithm. The original multi-signals are concatenated by using part of the information at the head and tail of each dimension or variate, and the IMFs of concatenated signal are deconcatenated by removing the transitions.
\section{Experimental results}
\label{sec_04}

A simulation dataset, as well as a real-world face image database, have been used to test and validate the proposed approach in order to show some of its potential applications. Several approaches are discussed: MVEMD, GiT-BEMD, standard one-dimensional EMD, EEMD, CEEMDAN and their corresponding proposed serial-versions. For the standard one-dimensional EMD and its variants, the input data was treated as a collection of one-dimensional slices, which enabled them to decompose the multi-signals. For the serial-EMD and its variants, on the other hand, such an operation is not necessary, as the input data has been naturally serialized to one dimension. The quality of IMFs and the time of decomposition are the two main performance indicators that have been used to evaluate these algorithms.

In the following subsections, we will detail the results under several situations (artificial multi-variate signals, artificial multi-dimensional images and face images). Both standard-version and serial-version EMD, EEMD and CEEMDAN were applied to these three situations, while MVEMD was only applied on artificial multi-variate signals and GiT-BEMD was only applied on artificial images and face datasets. For noise-assisted EMD algorithms, such as EEMD, CEEMDAN and GiT-BEMD, the noise standard deviation (NSTD), number of realizations (NR), maximum number of sifting iterations (MaxIter) and other hyper-parameters %set the values recommended in their original papers
are set as the values recommended in their respective original papers\cite{Wu2009, Torres2011, Al-Baddai2016}. All the experiments have been carried out on the same laptop (Intel(R) i7-6700HQ @ 2.6GHz).

\subsection{Artificial multi-variate signals}

\begin{figure}[ht]
    \centering{}
    \includegraphics[width=0.8\textwidth]{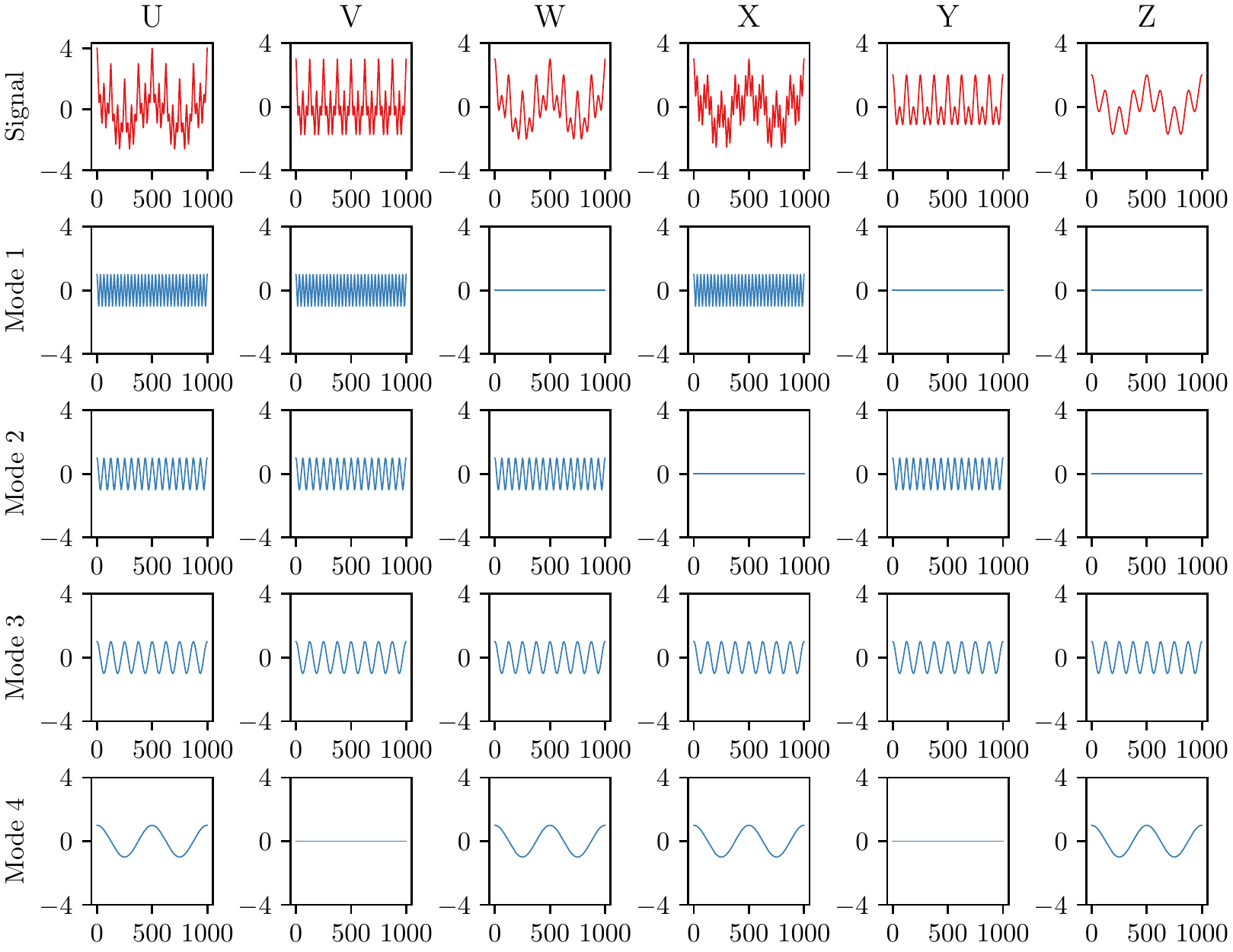}
    \caption{Synthetic multi-variate signals (U, V, W, X, Y, Z) with multiple frequency modes (32Hz, 16Hz, 8Hz, 2Hz). }
    \label{fig:C4}
\end{figure}

Similarly to MVEMD \cite{Rehman2010}, synthetic multi-variate time series with multiple frequency modes were considered. Each variate, shown in the top row of Figure \ref{fig:C4} (denoted by U, V, W, X, Y and Z), was constructed from a set of four sinusoids ($f_1=32Hz, f_2=16Hz, f_3=8Hz, f_4=2Hz$). One sinusoid was made common to all the components, whereas the remaining three sinusoidal components were combined so that the resulting signal had a common frequency mode ($f_3=8Hz$). The \emph{pick-up mask} eq. \ref{eq:pickup} is introduced to indicate which sinusoidal components will be included in each variate.

\begin{equation}\label{eq:pickup}
    mask = \begin{bmatrix}[l]
        1 & 1 & 0 & 1 & 0 & 0 \\
        1 & 1 & 1 & 0 & 1 & 0 \\
        1 & 1 & 1 & 1 & 1 & 1 \\
        1 & 0 & 1 & 1 & 0 & 1 \\
    \end{bmatrix}
\end{equation}

\begin{figure}[ht]
    \centering{}
    \includegraphics[width=0.8\textwidth]{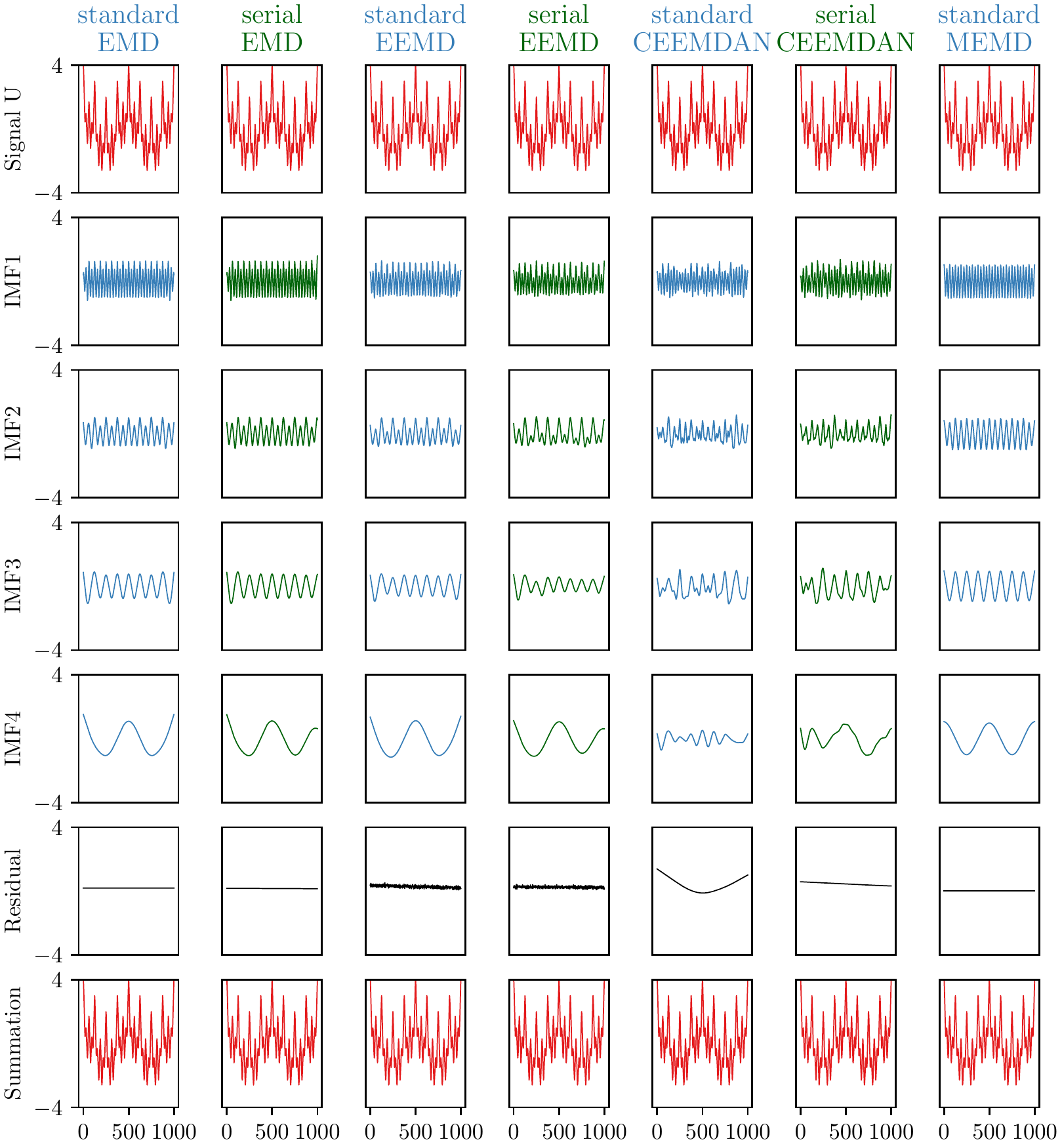}
    \caption{Decomposition results of one of the synthetic multi-variate signals (U) exhibiting multiple frequency modes (with $f_1=32Hz, f_2=16Hz, f_3=8Hz, f_4=2Hz$) via the proposed approach applied to other EMD algorithms. The IMFs extracted using serial-version algorithms (green) are similar to the IMFs extracted using standard-version algorithms (blue). Note that the subsequent IMFs (IMF5, IMF6...) are omitted since the amplitude and frequency of these IMF components are small enough to be ignored.}
    \label{fig:C5}
\end{figure}

The results for the IMFs for all algorithms are presented in Fig. \ref{fig:C5}. It can be observed that in terms of the amplitude and frequency of the IMFs, the resulting IMFs generated by serial-version algorithms are similar to the IMFs decomposed by standard-version algorithms, which indicates that the serialization of data has little effect on the standard EMD itself. Compared with standard-MVEMD, the amplitude and frequency of the IMFs extracted by standard-version algorithms and by serial-version algorithms are more inconsistent, which is known as the mode mixing problem. However, this is due to the defects of the EMD algorithm itself, not because of the introduction of serialization methods.

\begin{figure}[ht]
    \centering{}
    \includegraphics[width=0.8\textwidth]{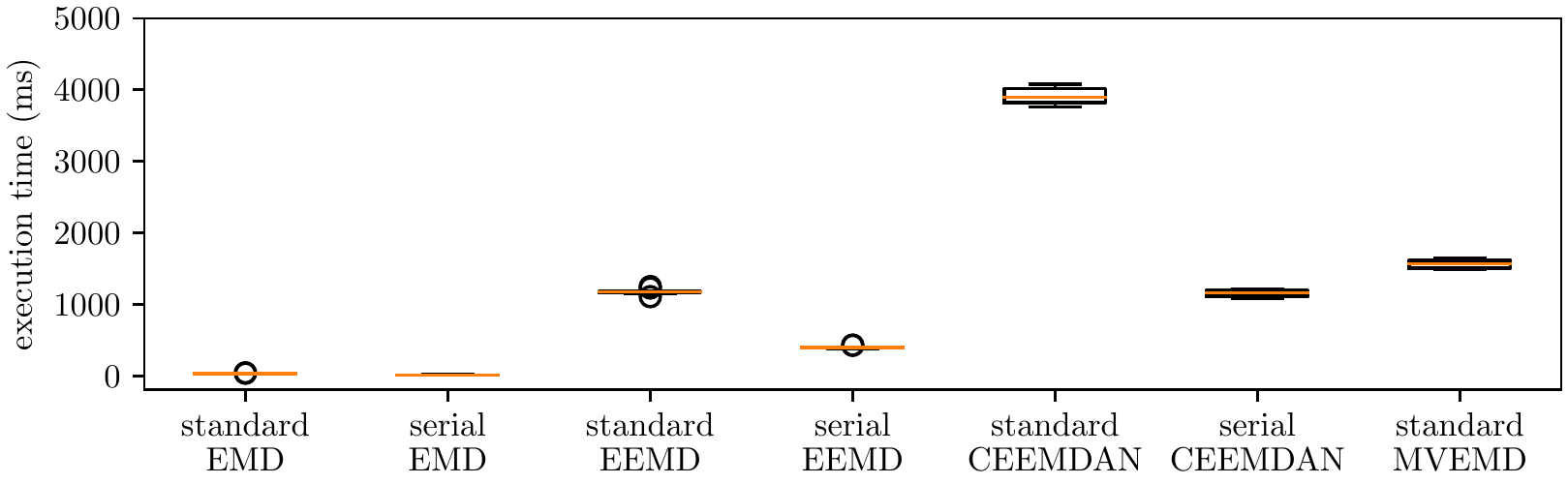}
    \caption{Comparison among various EMD algorithms for the artificial multi-variate signals (interval parameter, $D=50$) discussed in this paper in terms of the total time required. For each algorithm, the decomposition was repeatedly carried out 10 times. The black circles represent the outliers.}
    \label{fig:C6}
\end{figure}

Next, %the computational loads of these algorithms for synthetic multi-variate signals are investigated,
the computational load of these algorithms is investigated for synthetic multi-variate signals, as presented in Fig. \ref{fig:C6}. %Same with the following figures, 
In this figure, %I'M ASSUMING HERE THAT YOU WERE REFERRING TO FIGURE 6
%%%%% YES IT IS.
the bar ranges from Q1 (the first quartile) to Q3 (the third quartile) of the execution time distribution, the width represents the IQR (inter-quartile range), the median value is indicated by a red line across the bar and the outliers are shown in black circles. Compared with standard-MVEMD, the serial-version methods (EMD, EEMD, CEEMDAN) reduce the execution time by 99\%, 74\% and 26\%, respectively. Furthermore, the execution time of serial-CEEMDAN ($t=1154.97ms, std=46.32$), the most costly amongst the three serial-version algorithms, is less than that of standard-MVEMD ($t=1562.09ms, std=58.46$). For MVEMD, with the increase in the number of signals, the computational complexity shows %an increasing trend of geometric series,
a geometrically increasing trend, which makes it difficult to apply it on n-variate signals \cite{Tanaka2007}. For serial-EMD algorithms, the computational complexity can be reduced significantly, because it simplify the calculation of the mean envelope with the serialized one-dimensional signal.

\begin{figure}[ht]
    \centering{}
    \includegraphics[width=0.8\textwidth]{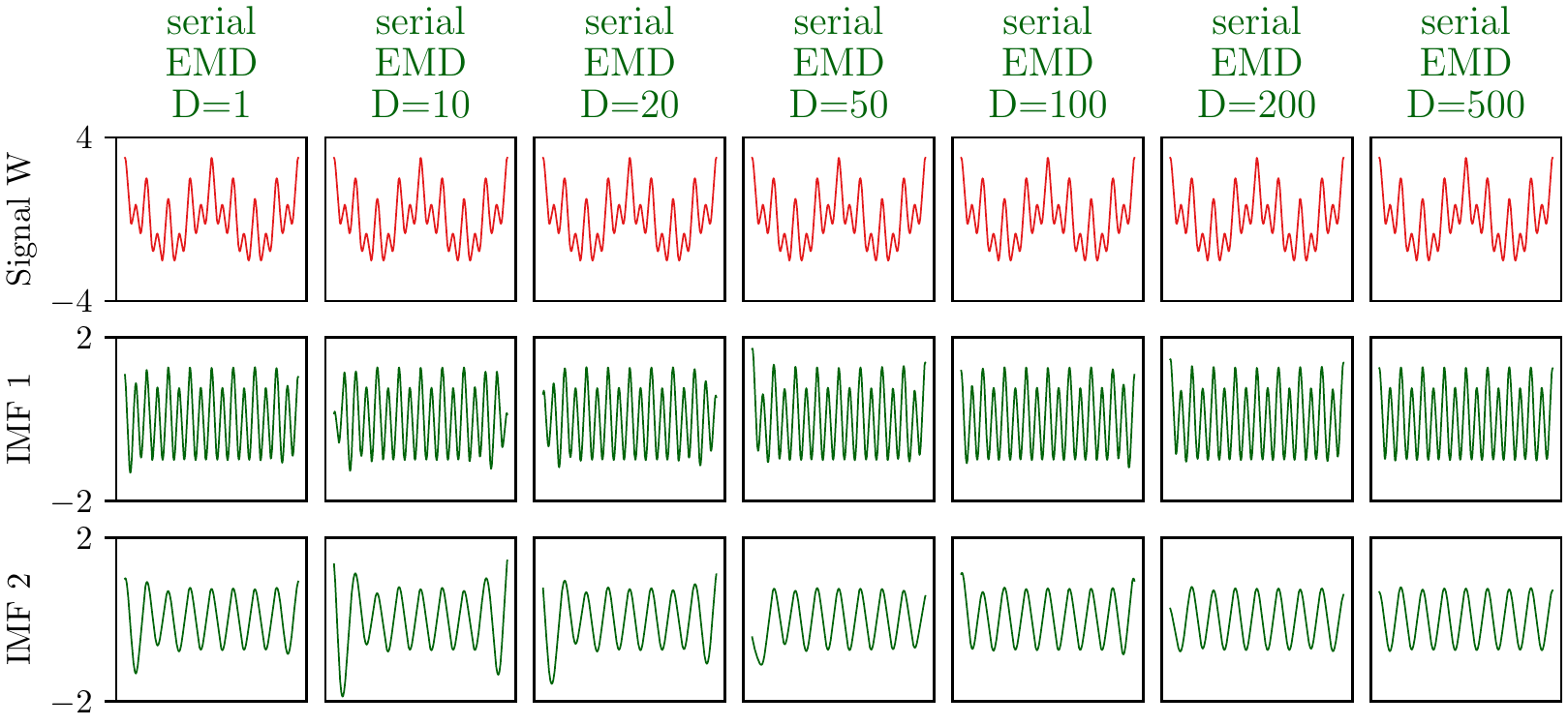}
    \caption{The components of IMFs obtained using serial-EMD with different interval parameters from $D=1$ (the lowest value for the interval parameter) to $D=500$ (half of the length of signals) are displayed. Only the first two IMFs are shown in this figure.}
    \label{fig:C7}
\end{figure}

Then, the effect of the interval values on the IMFs decomposed by different algorithms are compared. Fig. \ref{fig:C7} presents the results for several interval parameters $D$ from 1 (the lowest value for the interval) to 500 (half of the length of the signal). In the beginning (from $D=1$ to $D=50$), the larger the interval value is, the more similar the two ends of the IMF are to the middle part in terms of amplitude and frequency (for example, IMF2 in Fig. \ref{fig:C7}), which indicates that the transition part can be used to avoid the discontinuity in concatenated signals. However, for a larger interval (from $D=100$ to $D=500$), it is hard to distinguish between these IMFs, %which means there is no need to set over large interval between signals
which means that there is no need to set a large interval between the signals.

\begin{figure}[ht]
    \centering{}
    \includegraphics[width=0.8\textwidth]{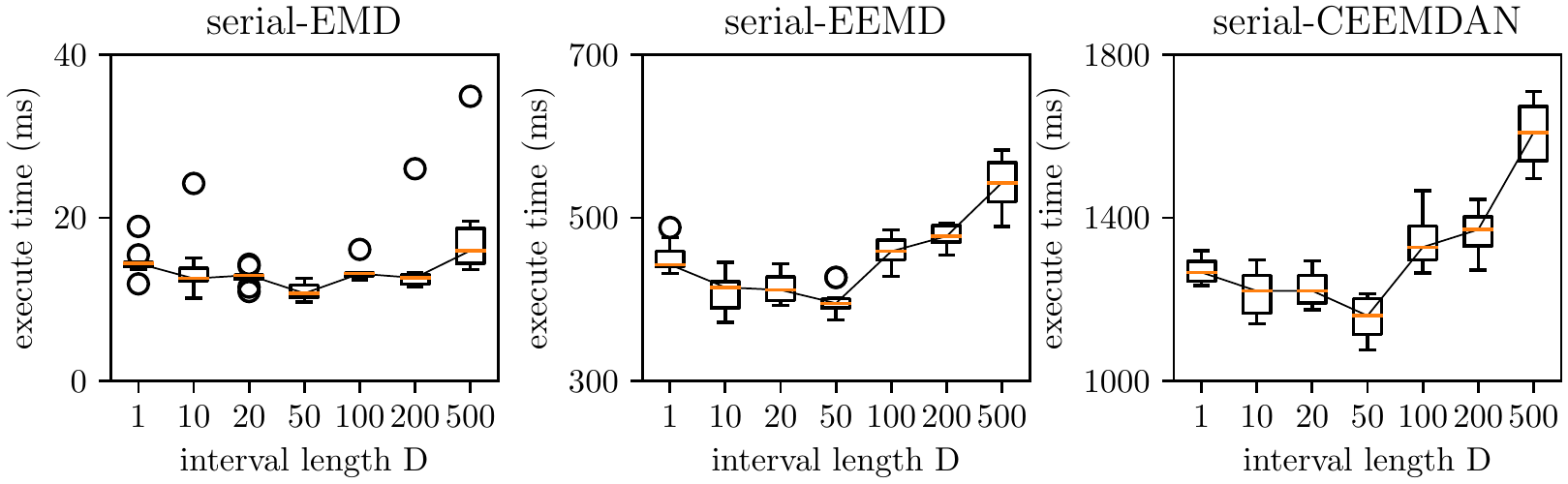}
    \caption{Comparison among various interval parameters from $D=1$ (the lowest value for the interval parameter) to $D=500$ (half of the length of signals) for serial-version algorithms in terms of the total time required. For each situation, the decomposition was repeatedly executed 10 times. The black circles represent the outliers.}
    \label{fig:C8}
\end{figure}

In addition, the effects of different interval values on the computational load of serial-EMD algorithms are also explored, as displayed in Fig. \ref {fig:C8}. From the beginning (from $D=1$ to $D=50$), the computational load decreases gradually and reaches its lowest value at $D=50$ (EMD: $t=10.96ms, std=0.92$, EEMD: $t=397.92ms, std=16.38$, CEEMDAN: $t=1154.97ms, std=46.32$). Increasing the interval value improves the continuity in concatenated signals, which speeds up the sifting process. However, for a larger interval (from $D=100$ to $D=500$), the computational load increases gradually, which indicates that the effect of the transition part reaches its bottleneck since the time cost is more related to the length of the signal.

In order to trade off between these two aspects, the hyper-parameter interval value $D$ should be set to 20\% of the length of the original signals. This value is also used in subsequent experiments.

\subsection{Artificial images}

\begin{figure}[ht]
    \centering{}
    \includegraphics[width=0.8\textwidth]{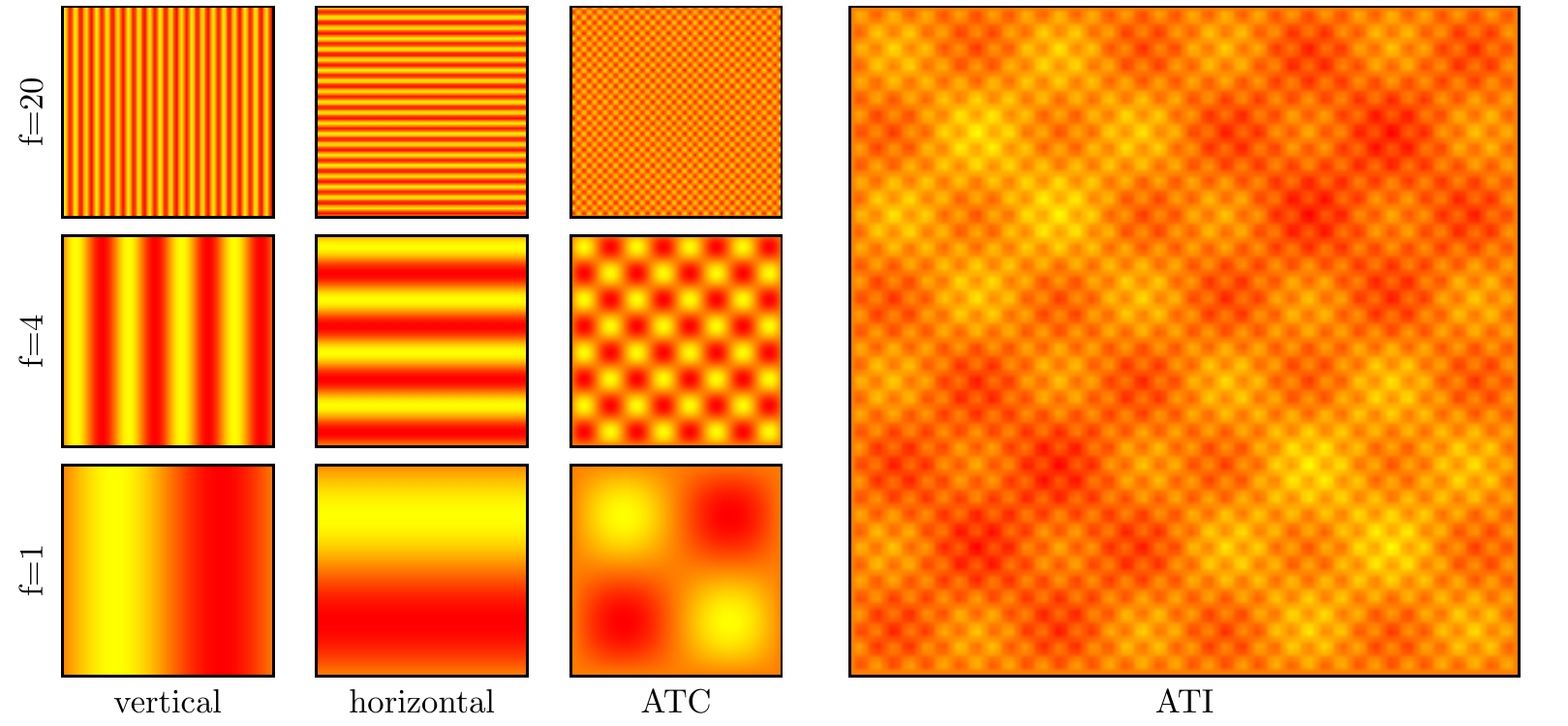}
    \caption{Left: Component results from a superposition of a horizontal ATI and a vertical ATI. Right: The original artificial texture image (ATI) produced by summing up the three ATC images.}
    \label{fig:C9}
\end{figure}

Similarly to MVEMD \cite{Al-Baddai2016}, an artificial texture image (ATI) of size $101 \times 101$ pixels is considered, which is composed of a superposition of artificial texture components (ATCs) of the same size. The ATCs represent harmonically varying spatial oscillations (sinusoids) with horizontal and vertical spatial frequencies ($f1=20$, $f2=4$, $f3=1$) and unit amplitudes, which are shown in Fig. \ref{fig:C9} on the left. The first ATC contains the highest spatial frequency, whereas the last ATC contains the lowest spatial frequency. These artificial textures provide a good performance indicator of the algorithm, even though some imperfections cannot be avoided. The ATI, the summation of these three ATCs, is shown in Fig. \ref{fig:C9} right.

\begin{figure}[ht]
    \centering{}
    \includegraphics[width=0.8\textwidth]{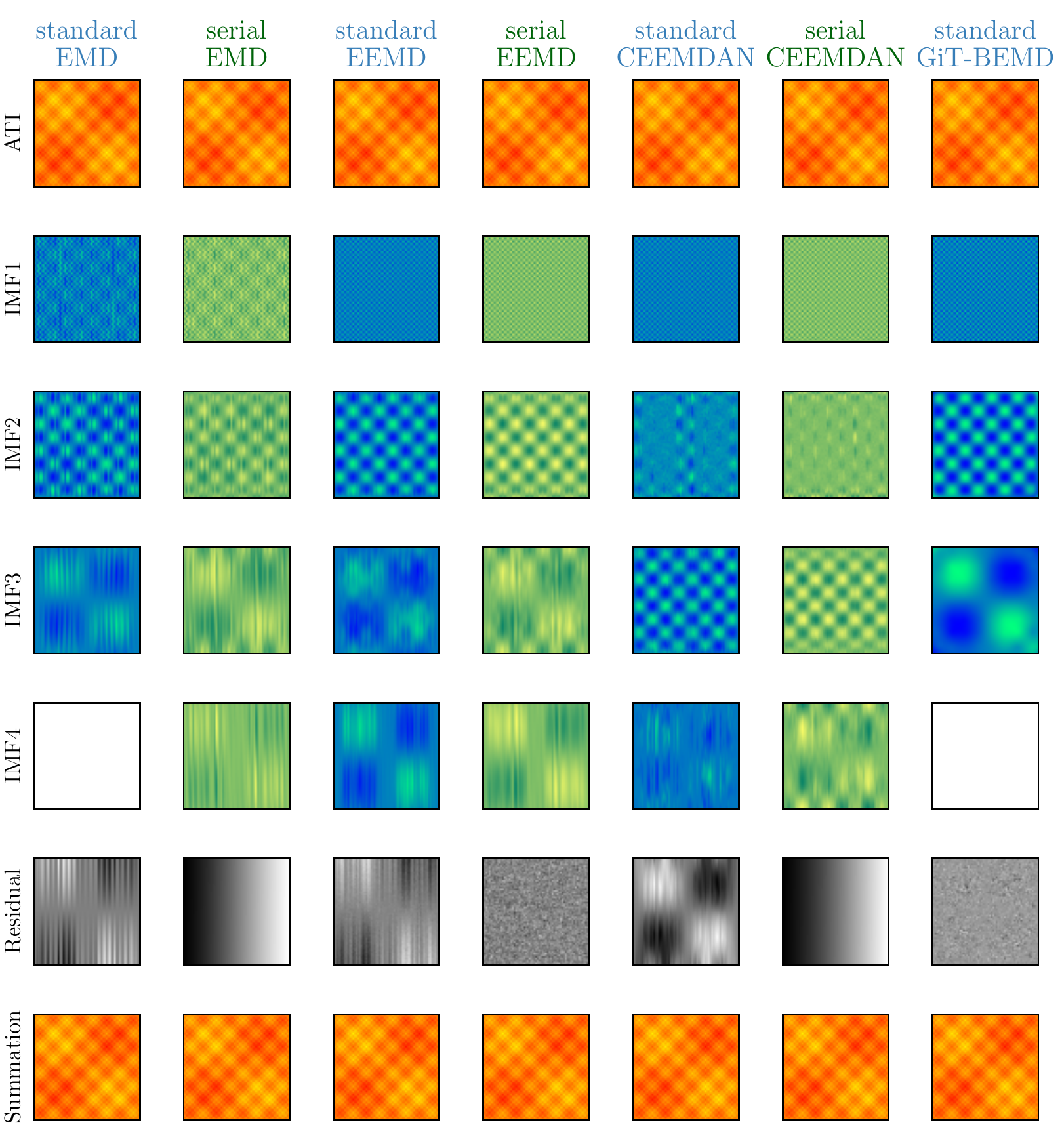}
    \caption{Decomposition of the ATI using standard-version algorithms and serial-version algorithms (interval parameter, $D=20$). Note that the subsequent IMFs (IMF4, IMF5...) are omitted since the amplitude and frequency of these IMF components are small enough to be ignored.}
    \label{fig:10}
\end{figure}

Firstly, the IMFs decomposed using each algorithm are investigated. Based on the former result of the artificial multi-variate signals, the optimal value of the interval parameter should be set at 20\% of the length of the original signal. Thus, considering that the size of the ATC is $101 \times 101$, the interval parameter should be set as 20. The resulting IMFs are displayed in Fig. \ref{fig:10}. Similarly to the previous experiment, the resulting IMFs generated by serial-version algorithms are similar to the IMFs decomposed by standard-version algorithms. For standard-version and serial-version algorithms, one single ATC can be decomposed into two or more IMFs (for example, some components of the second ATC appear in IMF2, IMF3 and IMF4 for the serial-EEMD algorithm), %which caused by the introduced artifacts during sifting process
which are caused by the artifacts introduced during the sifting process. Note that even though the IMFs decomposed using the EMD, EEMD and CEEMDAN algorithms contain some striated texture %compared with its decomposed by GiT-BEMD, it is more important to keep the characters of the original algorithms for serial-version algorithms. PLEASE CHECK IF WHAT I WROTE DOWN BELOW IS WHAT YOU WERE TRYING TO SAY WITH THIS SENTENCE:
%%%%% YES AND IT IS BETTER WITH YOUR VERSION.
compared with those decomposed using GiT-BEMD, the serial-version algorithms keep the same characteristics of the original ones.

\begin{figure}[ht]
    \centering{}
    \includegraphics[width=0.8\textwidth]{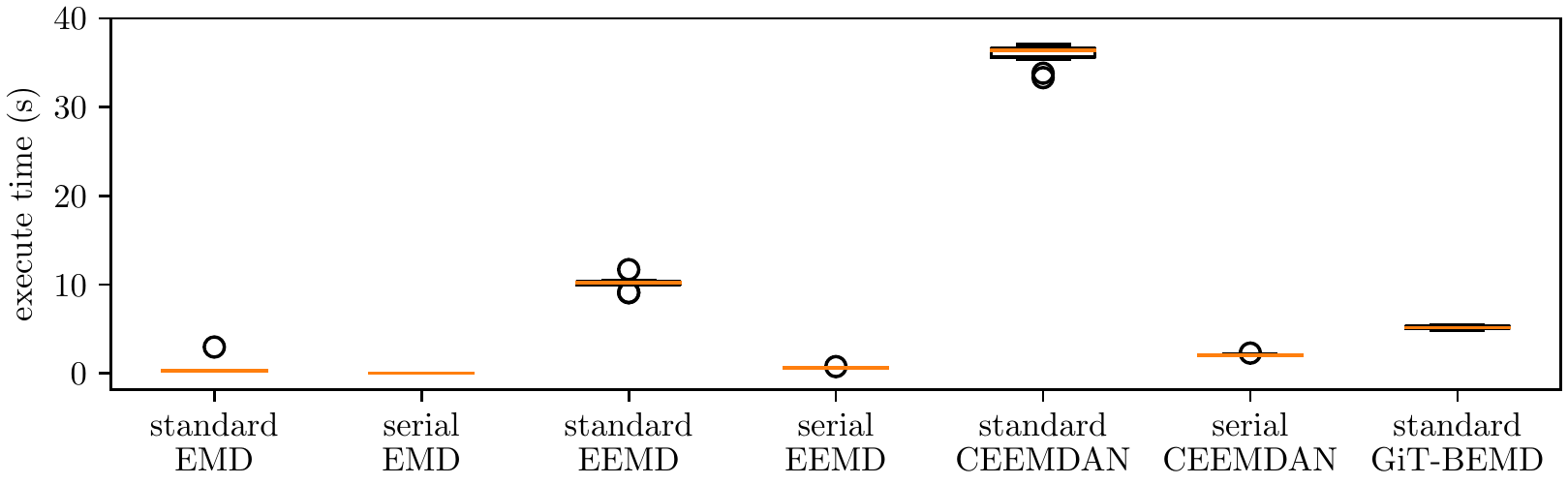}
    \caption{Comparison among various EMD algorithms for the ATI discussed in this paper in terms of the total time required. For each situation, the decomposition was repeatedly executed 10 times. The black circles represent the outliers.}
    \label{fig:C11}
\end{figure}

Subsequently, the computational loads of theses algorithms for ATI decomposition are explored, as presented in Fig. \ref{fig:C11}. When compared with standard GiT-BEMD, the serial-version methods (EMD, EEMD, CEEMDAN) reduce the execution time by 99\%, 87\% and 60\%, respectively. Moreover, the execution time of serial-EEMD ($t=6.41s, std=0.04$) is 1.5 times faster than that of standard-EEMD ($t=10.15s, std=0.69$). In addition, the execution time of serial-CEEMDAN ($t=20.58s, std=0.09$), the most costly amongst the three serial-version algorithms, is less than that of standard GiT-BEMD ($t=51.59s, std=0.19$). All these results clearly show that the serial-version algorithms are better than the standard-version algorithms in terms of computational load.

\subsection{Face database}

%Similarly to GiT-BEMD \cite{Al-Baddai2020}, real-world face images are also considered. 
To demonstrate the validity of the serialization method, real-world face images as the ones used with GiT-BEMD in \cite{Al-Baddai2020} were considered. The face database (AT \& T database) contains ten different images for each of the forty subjects in the database, which represents a total of 400 different images. All images were taken against a dark background, with the subjects in a frontal facial pose and with a varying orientation of the faces. All images are in a gray scale of 256 values, with an original size of $92 \times 112$ pixels.

\begin{figure}[ht]
    \centering{}
    \includegraphics[width=0.8\textwidth]{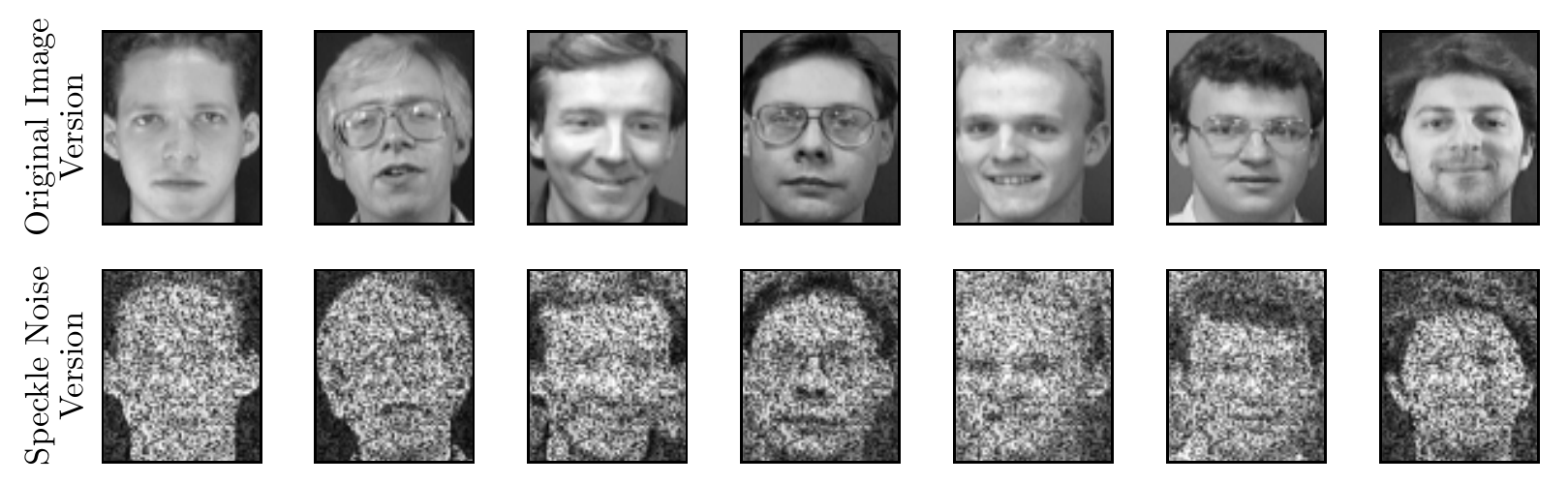}
    \caption{Top: %Original images with different person in AT & T face database. 
    Original images of different people from an AT \& T face database. Bottom: Corresponding noisy images with \emph{Speckle} noises at $SNR=-6$ dB.}
    \label{fig:C12}
\end{figure}

%The authors from \cite{Al-Baddai2020} reported that these face imagescorrupted by \emph{Speckle} noise with $SNR=-6$dB performed the worst (average accuracy=90\%) for face recognition scenario by using IMFs extracted by GiT-BEMD. 
The authors from \cite{Al-Baddai2020} reported that the worst result of the face recognition system was in the case when face images were corrupted by \emph{Speckle} noise with $SNR=-6$ dB. The classification system used IMFs extracted by GiT-BEMD.
The \emph{Speckle} noise adds multiplicative noise to the image elements $I(x,y)$ in the form $I(x,y)=I(x,y)+I(x,y)\hat{A} \cdot N(x,y)$, where $N(x,y)$ is a uniformly distributed zero-mean random noise with standard deviation $\sigma$, as provided in eq. \ref{eq:std_theta}. The presence of \emph{Speckle} noise in the images clearly has an effect on the edges and fine details which limit the resolution contrast. The original images and their corresponding noisy images are shown in Fig. \ref{fig:C12}. Note that only images with $SNR=-6$ dB \emph{Speckle} noise are used in the subsequent experiments.

\begin{equation}\label{eq:std_theta}
    f_U(n)=\left\{
    \begin{array}{rcl}
        1 / \sqrt{2\sigma} &  & for \quad n \in [-\sigma/\sqrt{2}, \sigma/\sqrt{2}] \\
        0                  &  & otherwise                                           \\
    \end{array}
    \right.
\end{equation}

\begin{figure}[ht]
    \centering{}
    \includegraphics[width=0.8\textwidth]{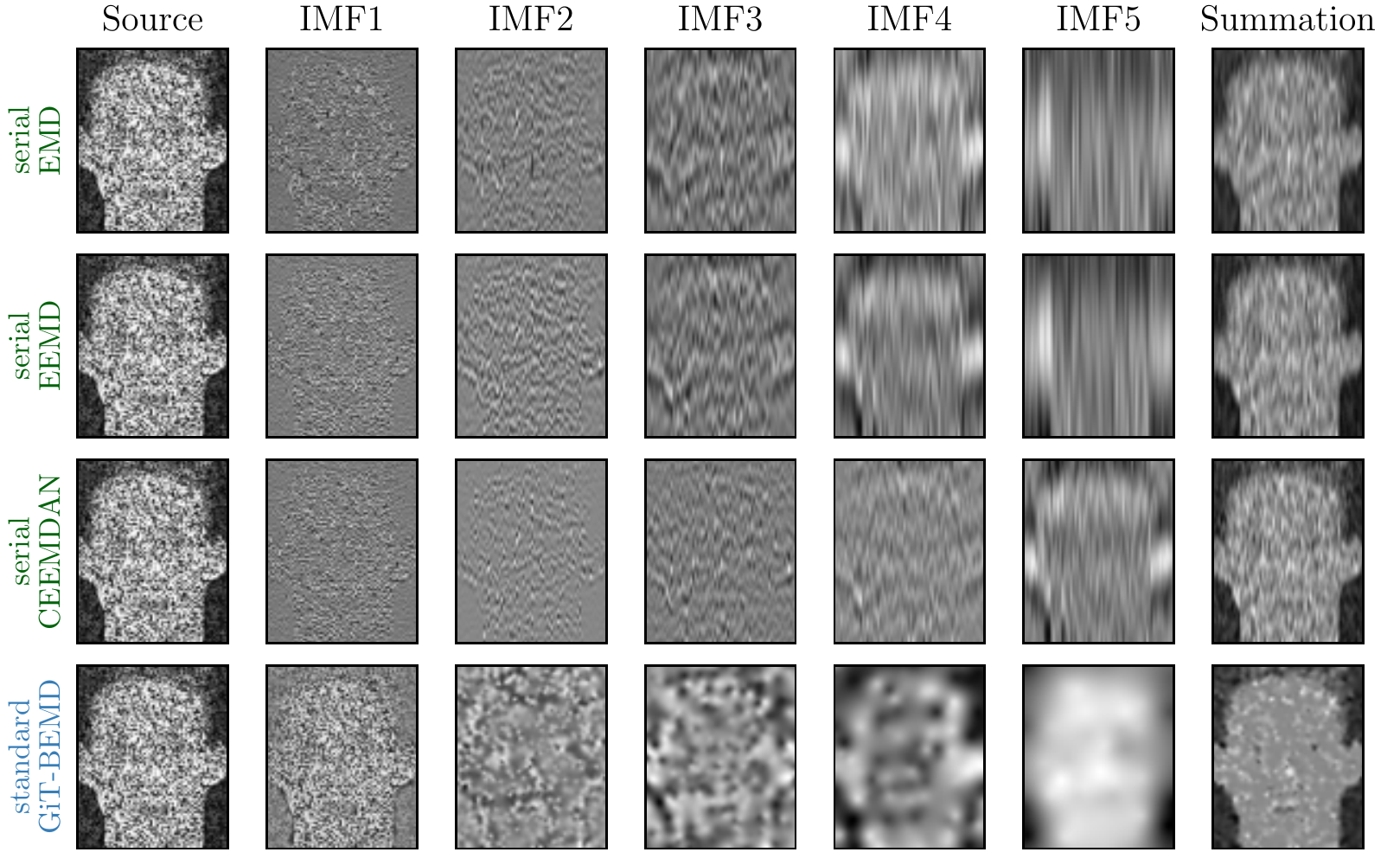}
    \caption{Examples of noisy images ($SNR=-6$ dB, \emph{Speckle} noise) and their corresponding IMFs through GiT-BEMD and serial-version algorithms. Note that the subsequent IMFs (IMF5, IMF6, ..., IMF12) are not shown in this figure. Summation column: summing up all the IMFs excluding the first two. %YOU ARE TALKING ABOUT EXCLUDING THE FIRST TWO IMFS, BUT IN FIGURE 15 YOU ARE ONLY EXCLUDING THE FIRST ONE. COULD YOU REVIEW WHICH ONE OF THESE IS CORRECT PLEASE? 
    %%%%% FOR HERE, THE SUMMATION COLUMN IS THE SUM OF ALL IMFS WITHOUT THE FIREST TWO IMFS. WHILE IN FIGURE 15, IT'S JUST A SKETCH TO SHOW HOW TO EXECUTE THE CLASSIFICATION WITH SEVERAL IMFS SO IT IS UNCERTAIN WHICH IMFS SHOULD BE EXCLUDED. LET'S KEEP BOTH SENTENCES IN HERE AND FIGURE 15.
    }
    \label{fig:C13}
\end{figure}

In Fig. \ref{fig:C13}, %the IMFs of one face noisy image decomposed by GiT-BEMD and serial-version algorithms315are presents
the resulting IMFs of the same noisy image of a face are presented when decomposed using both GiT-BEMD and serial-version algorithms. Considering that the size of the image is $92 \times 112$, the interval parameter should be set to 20. Similarly to the results for GiT-BEMD, the noise can also be separated from the original image using all the serial-version algorithms, and it is mostly captured by the first two IMFs. After removing the first few IMFs and summing up the remaining ones, the relatively clear images are presented in the summation column.

\begin{figure}[ht]
    \centering{}
    \includegraphics[width=0.8\textwidth]{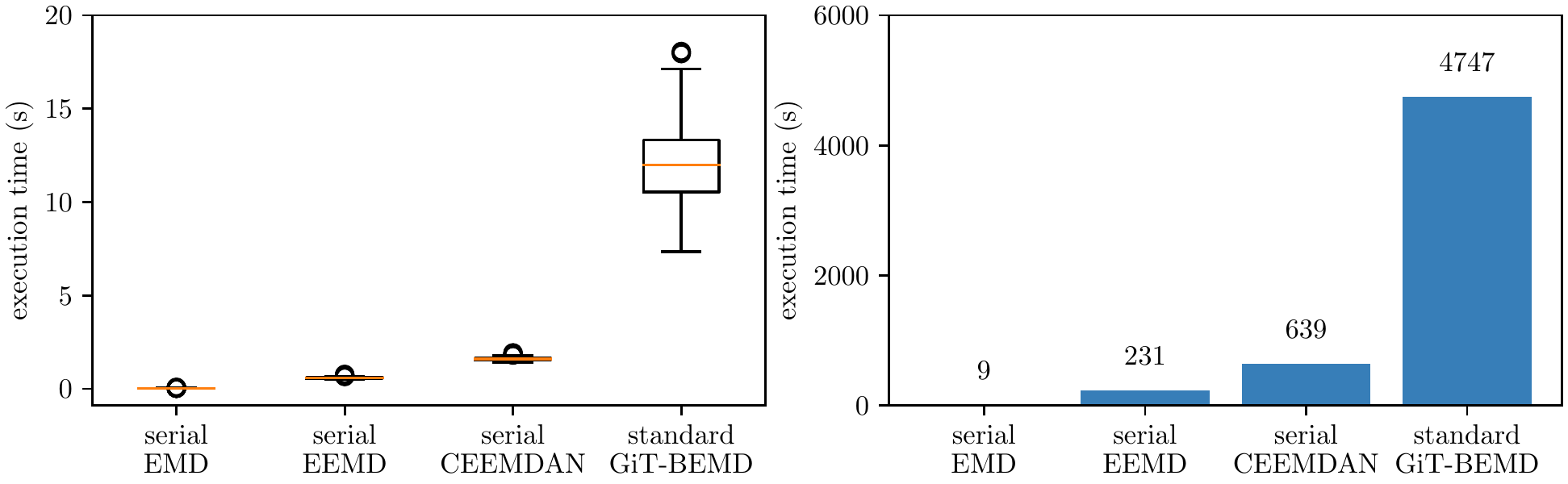}
    \caption{Left: Comparison between various algorithms for the processing of one single image in terms of the average time required. We calculated the average computational load per image. The black circles represent the outliers. Right: Comparison between various algorithms for the processing of the whole AT \& T database in terms of the total time required.}
    \label{fig:C14}
\end{figure}

The computational load of these algorithms for face images with \emph{Speckle} noise is also investigated, as displayed in Fig. \ref{fig:C14}. For all the serial-version algorithms, the average computational load per image (EMD: $t=0.024s, std=0.004$, EEMD: $t=0.578s, std=0.033$, CEEMDAN: $t=1.599ms, std=0.077$) is much less than that of GiT-BEMD ($t=11.86s, std=2.15$), as shown in Fig. \ref{fig:C14} (left). On top of that, the total execution times for the processing of the whole AT \& T database with noise are also presented in Fig. \ref{fig:C14} (right), which shows that the serial-version algorithms are greatly superior to GiT-BEMD when processing large amounts of data in terms of the execution time.

More specifically, %for decomposing the face AT & T database with 400 images, standard GiT-BEMD takes about 90 minutes, but serial-EMD, serial-EEMD, serial-CEEMDAN takes about 9 seconds, 4 minutes and 10 minutes, which reduces 99%, 95%, 88% execution time, respectively.
to decompose the face AT \& T database, which contains 400 images, standard GiT-BEMD takes about 90 minutes, while serial-EMD, serial-EEMD, serial-CEEMDAN take about 9 seconds, 4 minutes and 10 minutes instead, which reduces the execution time by 99\%, 95\% 88\%, respectively.

\begin{figure}[ht]
    \centering{}
    \includegraphics[width=0.8\textwidth]{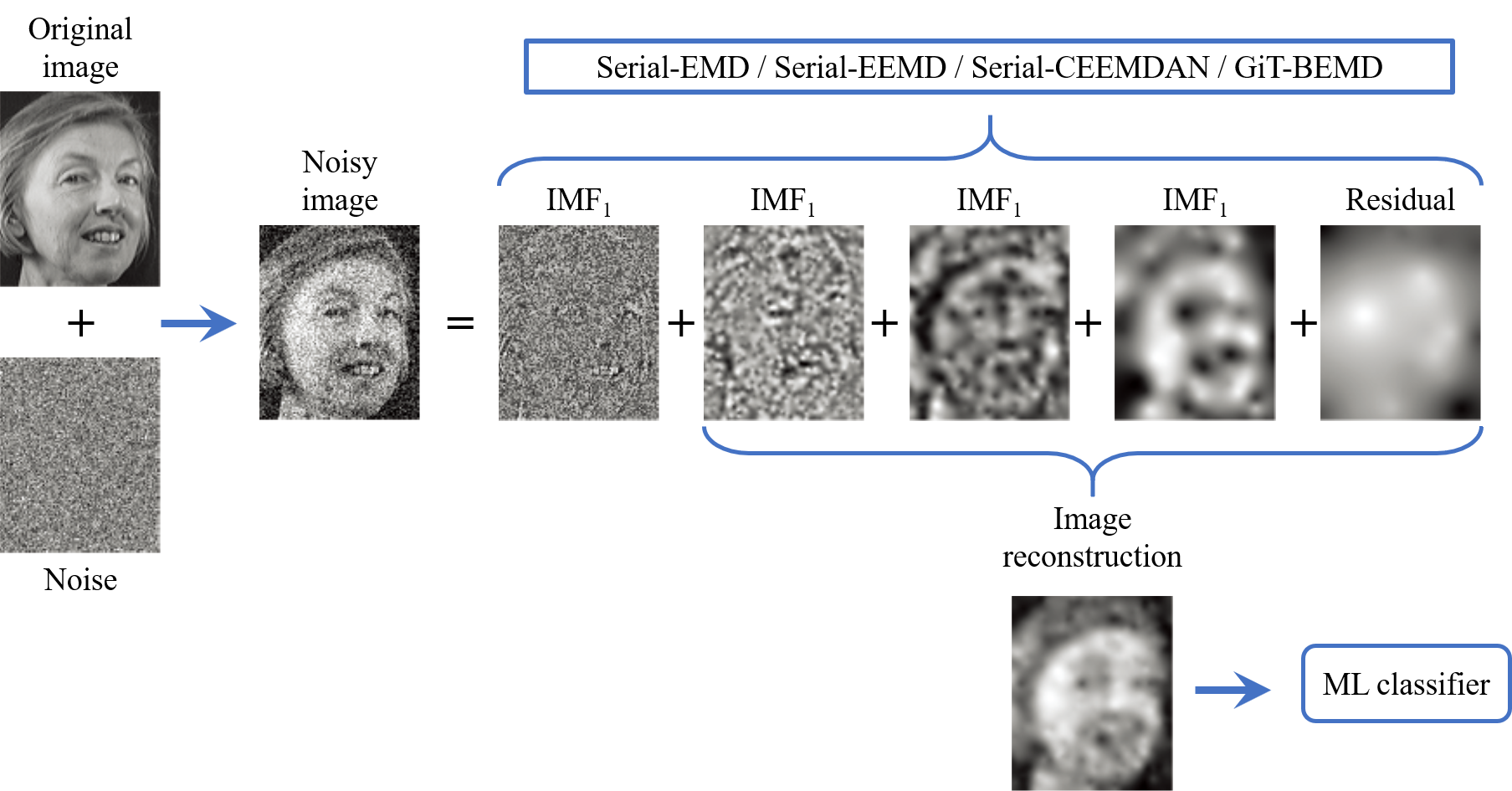}
    \caption{Schema of the face recognition experiment. All the IMFs except the first few are summed-up together to feed the classifier. When compared with the original experimental paradigm \cite{Al-Baddai2020}, our paradigm employed the summed-up way to process these IMFs instead of concatenating them.}
    \label{fig:C15}
\end{figure}

To explore %the ability of serial-version algorithms to capture noise
the ability to capture noise of the serial-version algorithms, the similar experimental paradigm as in \cite{Al-Baddai2020} is used, which is slightly different to the original paradigm to reduce the dimensions of the feature space. 
%Since serial-version algorithms can capture noise in the face image, therefore, either eliminating the first few IMFs or applying further filter to these IMFs, is reasonable to improve face recognition performance.
For image data with noise, its first few IMFs contain the high-frequency components, and its last few IMFs contain the low-frequency components. Since the \emph{Speckle} noise may be captured in the first few IMFs, a reasonable way to improve the face recognition performance is by eliminating them or applying further filters to these IMFs. The schema of this experiment is shown in Fig. \ref{fig:C15}. Images which have been corrupted by \emph{Speckle} noise at $SNR = -6$dB are first decomposed by serial-version algorithms. Since the first few extracted modes capture almost all of the noise contained in the image, these modes need to be removed. A new data vector is created by using KNN as the classification approach. All the experimental conditions, such as classification strategy, cross-validation strategy and its hyper-parameters, are consistent with those described in paper \cite{Al-Baddai2020}, except for the algorithm used for image decomposition and the number of summed-up IMFs.

\begin{figure}[ht]
    \centering{}
    \includegraphics[width=0.8\textwidth]{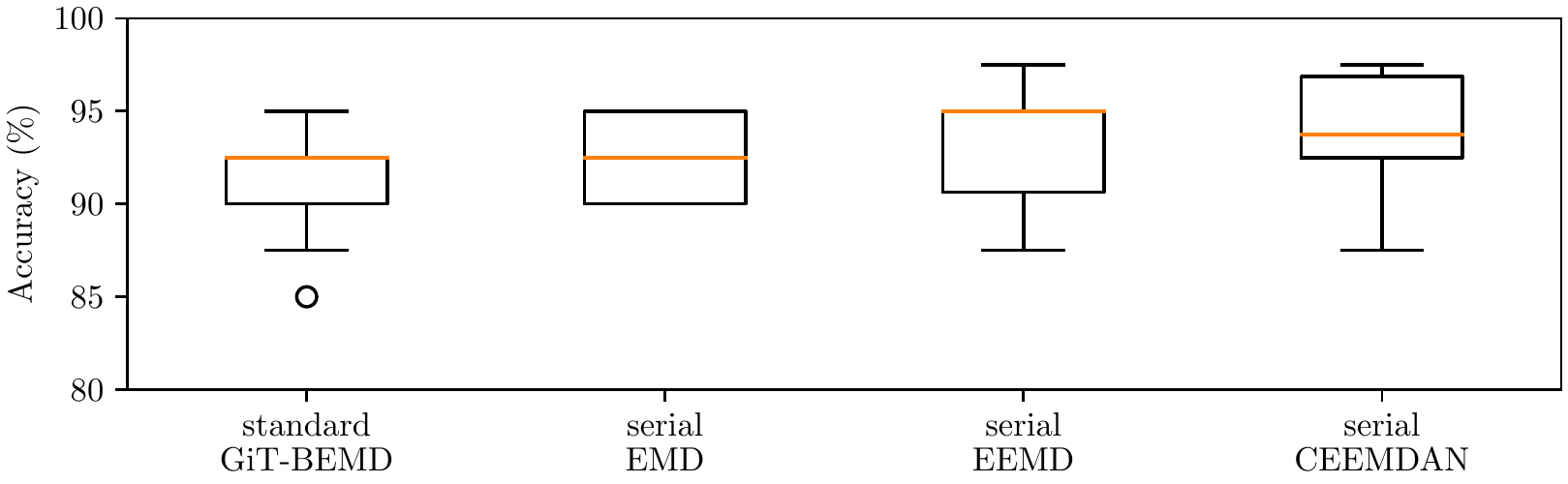}
    \caption{Highest face recognition accuracies for the standard GiT-BEMD and the proposed serial-version approaches by 10-fold cross-validation. Notice that the picked-up IMFs used for classification are different in each method. To achieve the highest accuracy, we summed up the 2nd to 5th IMFs, the 2nd to 7th IMFs, the 3rd to 6th IMFs and the 5th to 8th IMFs, respectively, which correspond to the red rectangle in Fig. \ref{fig:C17}.}
    \label{fig:C16}
\end{figure}

Face recognition performance results for the standard GiT-BEMD and for the proposed serial-version algorithms are presented in Fig. \ref{fig:C16}. In this case, several IMFs are summed up and flattened to reform a new one-dimensional vector to feed the KNN classifier. With 10-fold cross-validation, % compare with GiT-BEMD 
it can be seen that when compared with GiT-BEMD (accuracy=91.25\%, std=4.63), serial-EMD (accuracy=92.50\%, std=3.23), serial-EEMD (accuracy=93.24\%, std=3.91) and serial-CEEMDAN (accuracy=93.49\%, std=3.01) can all achieve a higher accuracy. 

\begin{table}[ht]
    \caption{The distribution of the number of IMFs decomposed by serial-version algorithms}
    \label{tbl:imf_distribution}
    \begin{center}
        \begin{tabular}{p{100pt}|p{50pt}|p{50pt}|p{50pt}|p{50pt}}
            \hline
            number of IMFs & 9 & 10 & 11 & 12 \\
            \hline
            number of trials & 66 & 960 & 172 & 2 \\
            \hline
        \end{tabular}
    \end{center}
\end{table}

For one single image, the number of IMFs decomposed by the three serial-version algorithms (EMD, EEMD, CEEMDAN) may be inconsistent. For different images, the number of IMFs decomposed by one algorithm may also change. Therefore, table \ref{tbl:imf_distribution} 
%shows the distribution of the number of IMFs decomposed by the three serial-version algorithms 400 images (1200 total trials). 
shows the distribution of the number of IMFs obtained when decomposing the 400 images (1200 total trials) using the three serial-version algorithms. Since the image can be decomposed into 10 IMFs in most of the trials, the maximum number of IMFs is set to 10 in subsequent experiments. For those images with less than 10 IMFs, we use zeros to fill the last IMF. If the number of IMFs exceeds 10, then the last few are discarded until only 10 remain.

\begin{figure}[ht]
    \centering{}
    \includegraphics[width=0.8\textwidth]{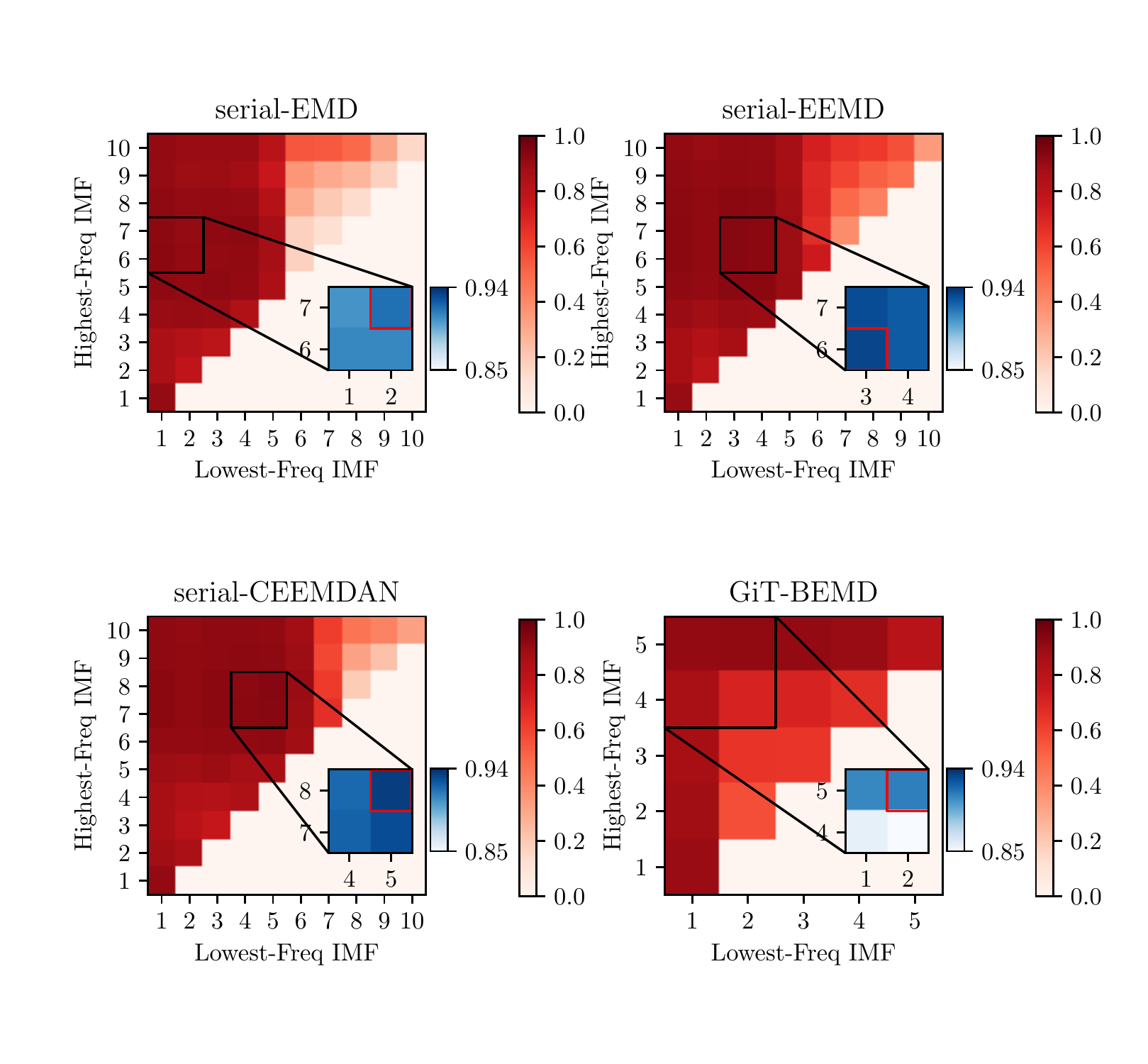}
    \caption{Accuracy heat map for all possible combinations of summed-up IMFs. The lowest-frequency IMF and the highest-frequency IMF refer to where to start picking up IMFs from and where to end the selection. The black square is a $2\times2$ regions which contains the highest accuracy. The topmost accuracy in each method is indicated with a red square in the sub-figure.
    %%%%% OK I HAVE UPDATED THE FIGURE.
    }
    \label{fig:C17}
\end{figure}

To investigate the power of the new approach in depth, all the experiments were carried out to test all the possible different ranges of IMFs. To do this, we use the highest/lowest-frequency, which are the first and last IMFs of the summed-up range, respectively. For this part, the range of highest-frequency IMF is from 1 to 10, which is the same as that of the lowest-frequency IMF. %PLEASE, COULD YOU QUICKLY DEFINE THE CONCEPTS OF HEAD AND TAIL IMF IN THE PAPER BEFORE YOU GO ON TO DESCRIBE THEIR RANGES? MAYBE SOMETHING LIKE "TO DO THIS, WE USE THE HEAD IMF AND TAIL IMF, WHICH ARE THE FIRST AND LAST IMFS OF THE RANGE (OF IMFS)"
%%%%% OK AND I HAVE ADDED THE CONCEPTS OF HEAD AND TAIL IMF AND I HAVE REPLACE THEM WITH HIGHEST/LOWEST-FREQUENCY IMF.
For example, when the highest-frequency IMF is set to 2 and the lowest-frequency IMF is set to 6, then all the IMFs from the 2nd IMF to the 6th IMF are selected to be summed up together and are fed to the KNN model. 
%The heat map 17 shows how the face recognition accuracy changes under different ranges of IMFs using different serial-version algorithms

Fig. \ref{fig:C17} shows the heat maps of the face recognition accuracies for different ranges of IMFs using the serial-version algorithms. In this figure, the darker the red, the higher the accuracy. The zoomed sub-figure of each heat map contains the best accuracy and shows the highest/lowest-frequency IMF corresponding to that highest accuracy.

Firstly, for serial-EMD, it can be observed that the highest accuracy is achieved with the combination of when the highest-frequency IMF is the 2nd IMF and the lowest-frequency is the 7th IMF (highest accuracy = 92.50\%). Likewise, for serial-EEMD (highest accuracy = 93.24\%) and serial-CEEMDAN (highest accuracy = 93.49\%), a similar result to serial-EMD can be observed, but with differently selected highest/lowest-frequency IMFs. When compared with the standard GiT-BEMD (highest accuracy = 91.25\%), all these serial-version methods can achieve a higher accuracy, which means that our approaches are better at separating noise from the synthetic images than GiT-BEMD.

Hence, it can be observed that the middle-range IMFs can achieve a higher accuracy than in other conditions, which shows that the IMFs in the middle range contain the most amount of information of the face images.

If we only use a single IMF to train the classifier, which corresponds to the diagonal in each heat map in Fig. 17, it can be seen that the first IMF achieves the highest accuracy out of any other IMF for all the algorithms. This indicates that the first IMF not only contains noise but also contains information of the original image. However, summing up multiple IMFs can achieve a higher accuracy, and note that in that case, the first IMF is not included in the ideal IMF range, which again may indicate that most of the information is kept in the middle range of the IMFs. 

In summary, our proposed method can not only achieve a higher accuracy than GiT-BEMD, but also has a great reduction in computational load in comparison, which shows that our methods are efficient in separating the noise from these synthetic face images and could be used in real-time applications.
\section{Conclusion}
\label{sec_05}

Empirical mode decomposing (EMD) represents a fully unsupervised data-driven technique that allows us to analyze non-linear and non-stationary data sets such as images, EMG signals and fMRI data. EMD encompasses a sifting process for estimating the intrinsic modes into which the data can be decomposed. In most of the existing implementations of EMD, %the sifting process affords the frequent estimation of
the sifting process allows for the frequency estimation of the upper and lower envelope surfaces interpolating extreme data points. The computation of these envelope surfaces becomes a problem in existing EMD variants due to its high computational load - the computation time grows exponentially with the data dimension, %which also caused the existing multi-EMD hardly expand to higher dimension.
which also causes the existing multi-EMD to be unsuitable for higher dimensions.

With the requirement of reducing the computational load during the sifting process for multi-variate or multi-dimensional EMD algorithms, we propose a novel approach, serial-EMD, based on the signal serialization technique. Firstly, the original multi-signals are concatenated using transitions, which are calculated using part of the information at the head and tail of each dimension or variate in the original signal. Then, after vectorization, the serialized one-dimensional signal is decomposed using EMD or its variants and IMFs for serialized signal are calculated. Finally, after removing the redundant transitions, the IMFs of the original multi-signals are extracted from the IMFs of the serialized signal by reshaping and slicing.

We have shown experimental results for both real and artificial signals in terms of the quality of IMFs and the computational load, thus demonstrating that serial-EMD can be applied to any multi-signal and for real-time applications. However, the serial-EMD is a double-edged sword: %as it concatenates multi-variate or multi-dimensional signals to one-dimensional signal, and uses various one-dimensional EMD algorithms to decompose it. THIS BIT IS NOT EXACTLY NECESSARY IN MY OPINION, AND IT COMPLICATES THE SENTENCE.
%%%%% OK, I AGREED WITH THAT :)
On the one hand, this algorithm only changes the shape of the data, so it is easy to combine with the existing EMD algorithms; on the other hand, since this method does not %involve the optimization of interpolation technology
modify the interpolation step, the resulting IMFs obtained by this algorithm are %difficult to get more optimized 
more difficult to optimize compared with other existing algorithms.

Altogether, we believe that serial-EMD offers a highly competitive alternative to existing multi-EMD algorithms and represents a promising technique for fast signal analysis.

\section{Acknowledgments}

This work was supported by the National Key R\&D Program of China (No. 2017YFE0129700), the National Natural Science Foundation of China (Key Program) (No. 11932013), the National Natural Science Foundation of China (No. 61673224), the Tianjin Natural Science Foundation for Distinguished Young Scholars (No. 18JCJQJC46100), and the Tianjin Science and Technology Plan Project (No. 18ZXJMTG00260).
\bibliographystyle{./elsarticle-num}
\bibliography{./main.bib}

\UseRawInputEncoding
\begin{thebibliography}{10}
\expandafter\ifx\csname url\endcsname\relax
  \def\url#1{\texttt{#1}}\fi
\expandafter\ifx\csname urlprefix\endcsname\relax\def\urlprefix{URL }\fi
\expandafter\ifx\csname href\endcsname\relax
  \def\href#1#2{#2} \def\path#1{#1}\fi

\bibitem{Huang1998}
N.~E. Huang, Z.~Shen, S.~R. Long, M.~C. Wu, H.~H. Snin, Q.~Zheng, N.~C. Yen,
  C.~C. Tung, H.~H. Liu, {The empirical mode decomposition and the Hubert
  spectrum for nonlinear and non-stationary time series analysis}\href
  {https://doi.org/10.1098/rspa.1998.0193} {\path{doi:10.1098/rspa.1998.0193}}.

\bibitem{Sharpley2006}
R.~C. Sharpley, V.~Vatchev, {Analysis of the intrinsic mode functions}\href
  {https://doi.org/10.1007/s00365-005-0603-z}
  {\path{doi:10.1007/s00365-005-0603-z}}.

\bibitem{Tolwinski2007}
S.~Tolwinski, {The Hilbert Transform and Empirical Mode Decomposition as Tools
  for Data Analysis Real Signals and the Hilbert Transform}.

\bibitem{Huang2005}
N.~E. Huang, N.~O. Attoh-Okine, {The Hilbert-Huang transform in engineering}.
\newblock \href {https://doi.org/10.1201/9781420027532}
  {\path{doi:10.1201/9781420027532}}.

\bibitem{Wu2009}
Z.~Wu, N.~E. Huang, {Ensemble empirical mode decomposition: A noise-assisted
  data analysis method}\href {https://doi.org/10.1142/S1793536909000047}
  {\path{doi:10.1142/S1793536909000047}}.

\bibitem{Torres2011}
M.~E. Torres, M.~A. Colominas, G.~Schlotthauer, P.~Flandrin, {A complete
  ensemble empirical mode decomposition with adaptive noise}, in: ICASSP, IEEE
  International Conference on Acoustics, Speech and Signal Processing -
  Proceedings.
\newblock \href {https://doi.org/10.1109/ICASSP.2011.5947265}
  {\path{doi:10.1109/ICASSP.2011.5947265}}.

\bibitem{Gallix2012}
A.~Gallix, J.~M. Górriz, J.~Ramírez, I.~A. Illán, E.~W. Lang, {On the
  empirical mode decomposition applied to the analysis of brain SPECT
  images}\href {https://doi.org/10.1016/j.eswa.2012.05.058}
  {\path{doi:10.1016/j.eswa.2012.05.058}}.

\bibitem{Navarro2012}
X.~Navarro, F.~Porée, G.~Carrault, {ECG removal in preterm EEG combining
  empirical mode decomposition and adaptive filtering}, in: ICASSP, IEEE
  International Conference on Acoustics, Speech and Signal Processing -
  Proceedings.
\newblock \href {https://doi.org/10.1109/ICASSP.2012.6287970}
  {\path{doi:10.1109/ICASSP.2012.6287970}}.

\bibitem{Celebi2012}
A.~T. Çelebi, S.~Ertürk, {Visual enhancement of underwater images using
  Empirical Mode Decomposition}\href
  {https://doi.org/10.1016/j.eswa.2011.07.077}
  {\path{doi:10.1016/j.eswa.2011.07.077}}.

\bibitem{Lei2011}
Y.~Lei, Z.~He, Y.~Zi, {EEMD method and WNN for fault diagnosis of locomotive
  roller bearings}\href {https://doi.org/10.1016/j.eswa.2010.12.095}
  {\path{doi:10.1016/j.eswa.2010.12.095}}.

\bibitem{UmairBinAltaf2007}
M.~{Umair Bin Altaf}, T.~Gautama, T.~Tanaka, D.~P. Mandic, {Rotation invariant
  complex empirical mode decomposition}, in: ICASSP, IEEE International
  Conference on Acoustics, Speech and Signal Processing - Proceedings.
\newblock \href {https://doi.org/10.1109/ICASSP.2007.366853}
  {\path{doi:10.1109/ICASSP.2007.366853}}.

\bibitem{Mandic2009}
D.~P. Mandic, V.~S.~L. Goh, {Complex Valued Nonlinear Adaptive Filters:
  Noncircularity, Widely Linear and Neural Models}.
\newblock \href {https://doi.org/10.1002/9780470742624}
  {\path{doi:10.1002/9780470742624}}.

\bibitem{Tanaka2007}
T.~Tanaka, D.~P. Mandic, {Complex empirical mode decomposition}\href
  {https://doi.org/10.1109/LSP.2006.882107}
  {\path{doi:10.1109/LSP.2006.882107}}.

\bibitem{Rilling2007}
G.~Rilling, P.~Flandrin, P.~Goncalves, J.~M. Lilly, {Bivariate empirical mode
  decomposition}\href {https://doi.org/10.1109/LSP.2007.904710}
  {\path{doi:10.1109/LSP.2007.904710}}.

\bibitem{UrRehman2010}
N.~{Ur Rehman}, D.~P. Mandic, {Empirical mode decomposition for trivariate
  signals}\href {https://doi.org/10.1109/TSP.2009.2033730}
  {\path{doi:10.1109/TSP.2009.2033730}}.

\bibitem{Rehman2010}
N.~Rehman, D.~P. Mandic, {Multivariate empirical mode decomposition}\href
  {https://doi.org/10.1098/rspa.2009.0502} {\path{doi:10.1098/rspa.2009.0502}}.

\bibitem{Wu2009a}
Z.~Wu, N.~E. Huang, X.~Chen, {The multi-dimensional ensemble empirical mode
  decomposition method}\href {https://doi.org/10.1142/S1793536909000187}
  {\path{doi:10.1142/S1793536909000187}}.

\bibitem{Nunes2003}
J.~C. Nunes, Y.~Bouaoune, E.~Delechelle, O.~Niang, P.~Bunel, {Image analysis by
  bidimensional empirical mode decomposition}\href
  {https://doi.org/10.1016/S0262-8856(03)00094-5}
  {\path{doi:10.1016/S0262-8856(03)00094-5}}.

\bibitem{Linderhed2009}
A.~Linderhed, {Image empirical mode decomposition: A new tool for image
  processing}\href {https://doi.org/10.1142/S1793536909000138}
  {\path{doi:10.1142/S1793536909000138}}.

\bibitem{Huang1996}
N.~E. Huang, Z.~Shen, S.~R. Long, M.~C. Wu, H.~H. Shih, N.-c. Yen, C.~C. Tung,
  H.~H. Liu, {The empirical mode decomposition and the Hilbert spectrum for
  nonlinear and non-stationary time series analysis}.

\bibitem{Bhuiyan2010}
S.~M. Bhuiyan, J.~F. Khan, R.~R. Adhami, {A novel approach of edge detection
  via a fast and adaptive bidimensional empirical mode decomposition
  method}\href {https://doi.org/10.1142/S1793536910000446}
  {\path{doi:10.1142/S1793536910000446}}.

\bibitem{Al-Baddai2016}
S.~Al-Baddai, K.~Al-Subari, A.~M. Tomé, J.~Solé-Casals, E.~W. Lang, {A
  green's function-based Bi-dimensional empirical mode decomposition}\href
  {https://doi.org/10.1016/j.ins.2016.01.089}
  {\path{doi:10.1016/j.ins.2016.01.089}}.

\bibitem{Hu2012}
X.~Hu, S.~Peng, W.~L. Hwang, {EMD revisited: A new understanding of the
  envelope and resolving the mode-mixing problem in AM-FM signals}\href
  {https://doi.org/10.1109/TSP.2011.2179650}
  {\path{doi:10.1109/TSP.2011.2179650}}.

\bibitem{Damerval2005}
C.~Damerval, S.~Meignen, V.~Perrier, {A fast algorithm for bidimensional
  EMD}\href {https://doi.org/10.1109/LSP.2005.855548}
  {\path{doi:10.1109/LSP.2005.855548}}.

\bibitem{Nunes2009}
J.~C. Nunes, �.~Deléchelle, {Empirical mode decomposition: Applications on
  signal and image processing}\href {https://doi.org/10.1142/S1793536909000059}
  {\path{doi:10.1142/S1793536909000059}}.

\bibitem{Liu2005}
Z.~Liu, S.~Peng, {Boundary processing of bidimensional EMD using texture
  synthesis}\href {https://doi.org/10.1109/LSP.2004.839700}
  {\path{doi:10.1109/LSP.2004.839700}}.

\bibitem{Wessel1998}
P.~Wessel, D.~Bercovici, {Interpolation with Splines in Tension: A Green's
  Function Approach}\href {https://doi.org/10.1023/A:1021713421882}
  {\path{doi:10.1023/A:1021713421882}}.

\bibitem{Al-Baddai2020}
S.~Al-Baddai, P.~Marti-Puig, E.~Gallego-Jutglà, K.~Al-Subari, A.~M. Tomé,
  B.~Ludwig, E.~W. Lang, J.~Solé-Casals, {A recognition–verification system
  for noisy faces based on an empirical mode decomposition with Green's
  functions}\href {https://doi.org/10.1007/s00500-019-04150-9}
  {\path{doi:10.1007/s00500-019-04150-9}}.

\bibitem{Yang2016}
Z.-J. Yang, X.~He, W.-Y. Xiong, X.-F. Nie, {Face Recognition under Varying
  Illumination Using Green's Functionbased Bidimensional Empirical Mode
  Decomposition and Gradientfaces}\href
  {https://doi.org/10.1051/itmconf/20160701015}
  {\path{doi:10.1051/itmconf/20160701015}}.

\end{thebibliography}

\end{document}